\documentclass[sn-mathphys]{sn-jnl}

\usepackage[utf8]{inputenc} 
\usepackage[T1]{fontenc}    
\usepackage{hyperref}       
\usepackage{url}            
\usepackage{booktabs}       
\usepackage{amsfonts}       
\usepackage{nicefrac}       
\usepackage{microtype}      
\usepackage{comment}        
\usepackage{mathtools}
\usepackage{xparse}
\usepackage{graphicx}
\usepackage{xcolor}
\usepackage{threeparttable, tablefootnote}
\usepackage{changepage}
\usepackage{bm}
\usepackage{subcaption}
\usepackage{amsmath}
\usepackage{array}
\usepackage{makecell}
\usepackage{adjustbox}

\jyear{2023}%

\theoremstyle{thmstyleone}%
\theoremstyle{thmstyletwo}%

\theoremstyle{thmstylethree}%

\begin{document}

\title[Geometry-aware Transformer]{GeoT: A Geometry-aware Transformer for Reliable Molecular Property Prediction and Chemically Interpretable Representation Learning}


\author[1]{\fnm{Bumju} \sur{Kwak}}\email{meson3241@gmail.com}

\author[2,3]{\fnm{Jiwon} \sur{Park}}\email{jwpark21@snu.ac.kr}
\author[4]{\fnm{Taewon} \sur{Kang}}\email{taewonkang@kaist.ac.kr}
\author*[5]{\fnm{Jeonghee} \sur{Jo}}\email{page1024@snu.ac.kr}
\author[6]{\fnm{Byunghan} \sur{Lee}}\email{bhlee@seoultech.ac.kr}

\author*[3,5,7,8]{\fnm{Sungroh} \sur{Yoon}}\email{sryoon@snu.ac.kr}

\affil[1]{\orgdiv{Recommendation Team}, \orgname{Kakao Corporation}, \orgaddress{\state{Gyeonggi}, \postcode{13529}, \country{Republic of Korea}}}

\affil[2]{ \orgname{LG Chem}, \orgaddress{\city{Seoul}, \postcode{07795}, \country{Republic of Korea}}}

\affil[3]{\orgdiv{Interdisciplinary Program in Artificial Intelligence}, \orgname{Seoul National University}, \orgaddress{\city{Seoul}, \postcode{08826}, \country{Republic of Korea}}}

\affil[4]{\orgdiv{Department of Materials Science and Engineering}, \orgname{Korea Advanced Institute of Science and Technology (KAIST)}, \orgaddress{\city{Daejeon}, \postcode{34141}, \country{Republic of Korea}}}

\affil*[5]{\orgdiv{Institute of New Media and Communications}, \orgname{Seoul National University}, \orgaddress{\city{Seoul}, \postcode{08826}, \country{Republic of Korea}}}

\affil[6]{\orgdiv{Department of Electronic Engineering}, \orgname{Seoul National University of Science and Technology}, \orgaddress{\city{Seoul}, \postcode{01811}, \country{Republic of Korea}}}

\affil*[7]{\orgdiv{Department of Electrical and Computer Engineering}, \orgname{Seoul National University}, \orgaddress{ \city{Seoul}, \postcode{08826}, \country{Republic of Korea}}}

\affil*[8]{\orgdiv{Artificial Intelligence Institute}, \orgname{Seoul National University}, \orgaddress{ \city{Seoul}, \postcode{08826}, \country{Republic of Korea}}}

\abstract{
In recent years, molecular representation learning has emerged as a key area of focus in various chemical tasks. However, many existing models fail to fully consider the geometric information of molecular structures, resulting in less intuitive representations. Moreover, the widely used message-passing mechanism is limited to provide the interpretation of experimental results from a chemical perspective. To address these challenges, we introduce a novel Transformer-based framework for molecular representation learning, named the Geometry-aware Transformer (GeoT). GeoT learns molecular graph structures through attention-based mechanisms specifically designed to offer reliable interpretability, as well as molecular property prediction. Consequently, GeoT can generate attention maps of interatomic relationships associated with training objectives. In addition, GeoT demonstrates comparable performance to MPNN-based models while achieving reduced computational complexity. Our comprehensive experiments, including an empirical simulation, reveal that GeoT effectively learns the chemical insights into molecular structures, bridging the gap between artificial intelligence and molecular sciences.}




\maketitle

Quantum mechanical calculations have been used in the development of chemicals in various fields, such as drugs and catalysts. Density functional theory (DFT) is the most widely used computational methods of quantum mechanics (QM) modeling \cite{mardirossian2017thirty}, but it requires a great deal of computation to predict the properties of even a small molecule.
For this reason, several machine learning-based techniques have been explored as cost-effective alternatives \cite{rupp2012fast, lorenz2004representing, behler2007generalized, bartok2010gaussian}. In particular, deep learning has been used to predict molecular properties including energy and forces \cite{duvenaud2015convolutional, schutt2017schnet, unke2019physnet, anderson2019cormorant, yoo2020graph, klicpera2020directional, choukroun2021geometric}. 

It is common to regard a molecule as a graph in which the atoms are nodes and the edges are bonds in a message passing neural network (MPNN) \cite{gilmer2017neural}. SchNet \cite{schutt2017schnet}, PhysNet \cite{unke2019physnet} and several other MPNNs constructed the localized messages those are centered on atoms by applying continuous filters based on interatomic distances.
More recent networks such as DimeNet \cite{klicpera2020directional}, DimeNet++ \cite{klicpera_dimenetpp_2020}, and GemNet \cite{gasteiger2021gemnet} explicitly incorporate angle computation to represent molecular conformations. However, these previous models use a cutoff distance, which restricts the receptive field of atom-based convolutions, resulting in computationally intensive operations. 

An MPNN's localized view has inherent limitations. From a chemical point of view, all atom-atom pairs should be considered as essential components in molecular property prediction models, regardless of their interatomic distances.
However, the localized convolutions of MPNN conflict with the conformational behavior of molecules, since MPNN with a finite cutoff value assumes that messages are transferred within a restricted region. The use of a fixed cutoff distance fails to capture long-range interatomic relationships, in which the actual interactions between all the pairs of atoms in a molecule are determined by charge and distance. 

To overcome those limitations, we adopt Transformer \cite{vaswani2017attention} as our model framework for molecular graphs. Transformer consists of self-attention blocks learning the relations between two components from the sequence. If a graph can be represented as a sequence without loss of its topological information, Transformer can be a viable alternative to MPNN. The self-attention mechanism in Transformers can learn relationships between entities regardless of their positions, whereas MPNNs are limited to localized neighbors due to their restricted receptive fields.

Since Transformers were initially developed for processing sequential data with a specific order, they are generally not well-suited for handling order-invariant graph data. To overcome this problem, several previous Transformer-based architectures \cite{maziarka2020molecule, yoo2020graph, rong2020self, ying2021transformers, chen2022structure} were proposed, however, these methods cannot fully consider geometric information, which is the key factor of describing the nature of molecules. For example, \cite{yoo2020graph, rong2020self, ying2021transformers} blindly encodes an atom-atom interaction as a categorized representation of bond type. There are two problems with this categorization. Firstly, categorization of bond types cannot consider its individual length information, which depends on associated atoms types. Secondly, it does not allow for the consideration of cases where atoms are not bonded but are closely located to each other. 
Moreover, \cite{yoo2020graph, rong2020self} require excessive prior knowledge such as valence or aromaticity for molecular property prediction, which can be a limitation for exploring little-known molecules.

\begin{figure} 
    \centering
    \includegraphics[width=0.96\textwidth]{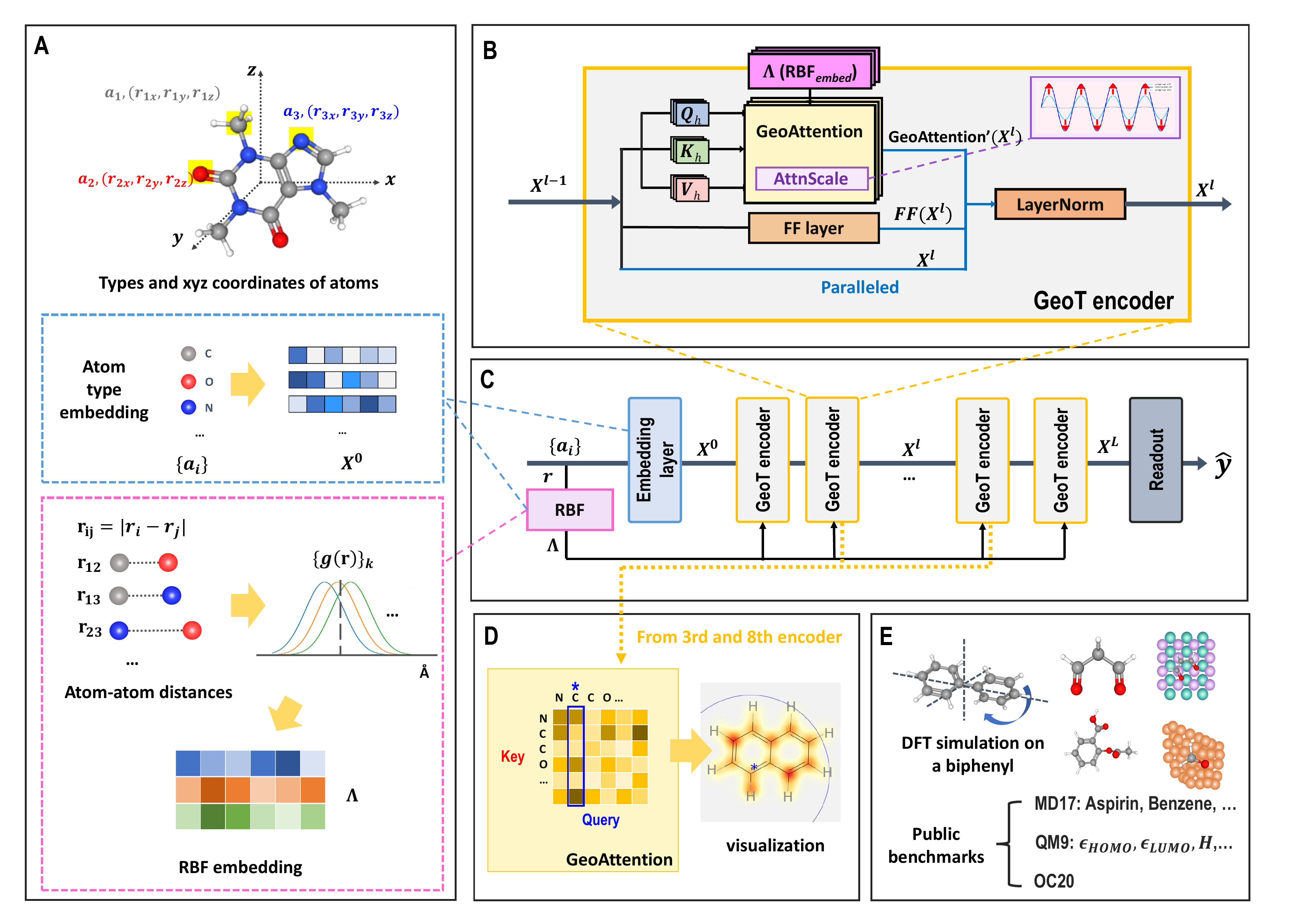}
\caption{Overview of Geometry-aware Transformer. \textbf{(A)} The embedding of a type of each atom and its position of a molecule as an input. A set of atom types $\{a_i\}$ are embedded as an inputs $X_0$ of GeoT in the learnable embedding layer. On the other hand, a set of atom-atom distances are represented as $k$ different RBFs ($\text{RBF}_{emb}$), to make $\boldsymbol{\Lambda}$ embedding matrix. Different from embedding layer for atom types, the distance embedding $\boldsymbol{\Lambda}$ from $\text{RBF}_{emb}$ is used an input of each GeoT encoder, which is a block of GeoT.  
\textbf{(B)} The internal structure of GeoT encoder accompanied with GeoAttention. The query $\boldsymbol{Q}$, key $\boldsymbol{K}$, and value $\boldsymbol{V}$ are obtained from the former layer $X^{l-1}$, and a matrix $\boldsymbol{\Lambda}$ are inputs of GeoAttention before applying the AttnScale. In addition, the components of each GeoT encoder are parallelized, which is different from the original self-attention. \textbf{(C)} The architecture of GeoT. It has one atom type embedding layer at the front, $L$ GeoT encoder blocks, and the readout layer at the end of the network. To extract attention feature maps for visualization, we used the 3rd and 8th GeoT encoders. \textbf{(D)} The example of attention map visualization from GeoAttention. Given a query atom (denoted as blue asterisk), associated attention weights of other atoms are shown as yellow-red shades. More red regions have higher attention weights associated with a given query atom. \textbf{(E)} The benchmarks used for model performance: MD17 and QM9 datasets consist of small-sized molecules, while OC20 dataset is a pair of surface and substrate. }
\label{fig:general_architecture}
\end{figure}

 To address the above issues, we aim to develop the model considering the nature of molecules represented by graphs with geometric information, named geometry-aware Transformer (GeoT). The overview of GeoT architecture is presented in Fig. \ref{fig:general_architecture}. 
 To achieve the interpretable prediction results, we introduce several modifications on the self-attention mechanism in GeoT. By incorporating these concerns, we propose two strategies into self-attention: 1) introducing $k$ radial basis function ($\text{RBF}_{emb}$) for embedding of geometric information of atom-atom distances and 2) replacing the softmax with the alternative scaling method to enhance more important atom features. We named our modified self-attention as GeoAttention.

The modifications described above are based on the intuition that different parts of a molecule contribute differently to various molecular properties. To encourage the model to attend to the most significant parts of a molecule, it is not desirable to use softmax, which indiscriminately adjusts all entities to the same scale.

We investigated the relationship between GeoT's attention pattern on different molecules and the associated chemical insights of the training objective to verify our intuition. Specifically, we investigated two contrasting molecular properties: $\epsilon_{LUMO}$ and $H$ cases, which are the energy of lowest unoccupied molecular orbital (LUMO) and molecular enthalpy, respectively. 
Surprisingly, we found salient pattern from GeoAttention, which is the modified self-attention in GeoT trained with $\epsilon_{LUMO}$.
Additionally, we conducted the DFT simulation on the case of biphenyl by predicting on its energy over the change of conformation, and verified superior generalizability of GeoT.

We also evaluate the prediction performance on the public benchmarks. First, we evaluated the prediction performance of GeoT on three public benchmark datasets MD17 \cite{chmiela2017machine, chmiela2019sgdml, chmiela2018towards}, QM9 \cite{ruddigkeit2012enumeration, ramakrishnan2014quantum}, and OC20 \cite{ocp_dataset}, which are the most widely used molecular property prediction tasks in current days. Second, we did an ablation study of the proposed model refinements to verify the effectiveness of GeoT. By comparing the results from those experiments, we concluded that our GeoAttention has additional advantages. 
 We summarize the contributions of GeoT as below.
\begin{itemize}

    \item Development of a geometry-aware Transformer architecture for molecular graphs, which incorporates geometric information of molecules for intuitive and interpretable representation learning.
    
    \item Introduction of the model refinement strategies enabling molecular graph representations to improve model performance and robustness.

    \item Verification of better interpretability and generalizability produced by the GeoAttention from a chemical perspective.

    \item Evaluation of the model performance over a range of benchmarks and the empirical study for molecular property prediction. 

\end{itemize}

\section{Result}
\label{Result}

\begin{figure} 
  \centering
  \includegraphics[width=0.96\textwidth]{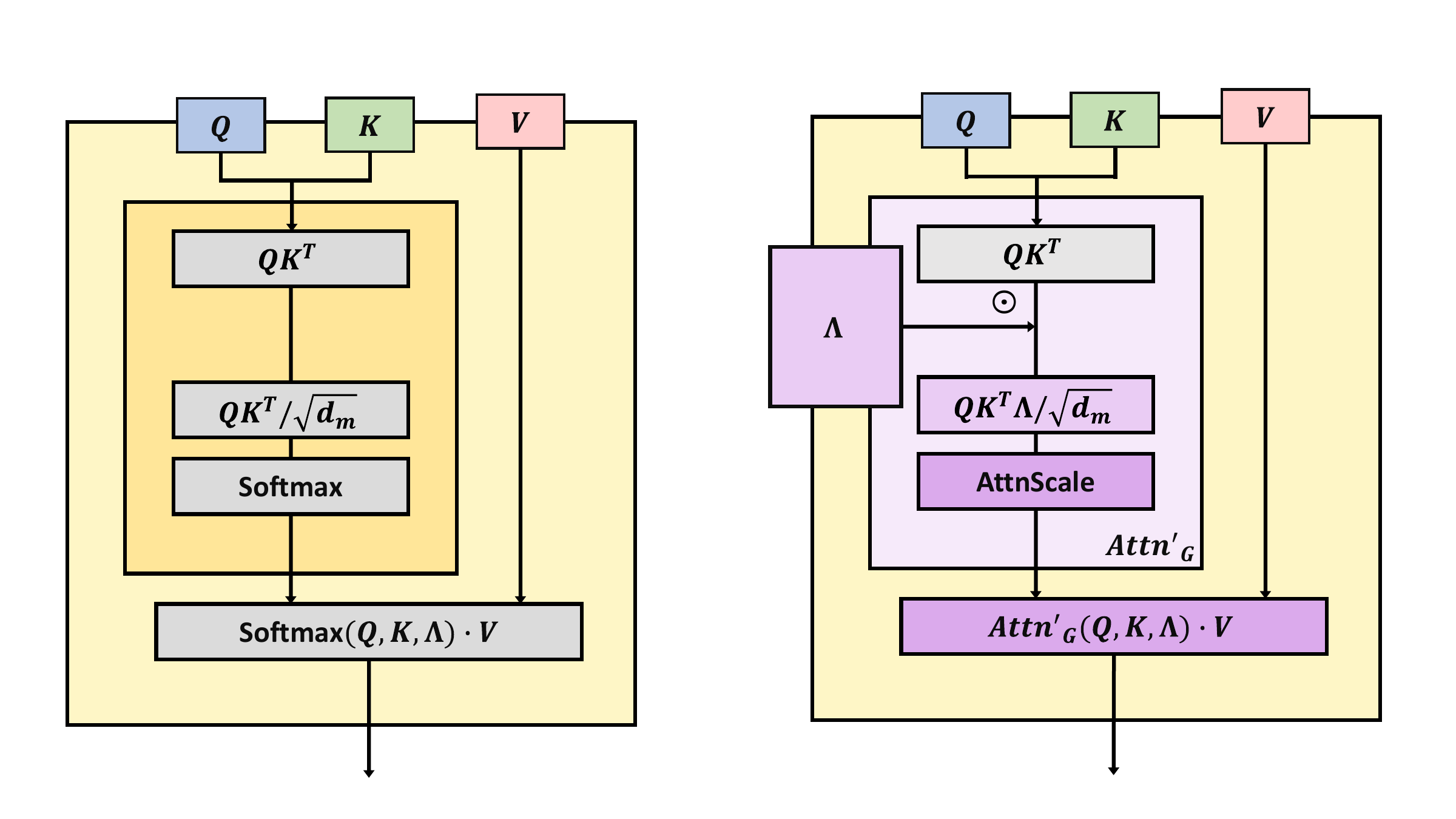}
  \caption[Comparison between the standard self-attention (left) and GeoAttention (right).]{Comparison between the standard self-attention (left) and GeoAttention (right). The key differences of two architectures are that the GeoAttention 1) used the $\mathrm{RBF}_{emb}$ to represent distances of atom pairs, and 2) removed softmax function existed in the standard self-attention.}
  \label{fig:diff_architecture}
\end{figure}

We analyzed the performance of GeoT in both qualitative and quantitative manner. For qualitative evaluation of the interpretability of GeoT, we visualize the trained attention score of GeoAttention in various cases. Visualization of the weights in self-attention has widely been used to provide the semantic relationships between different elements of data \cite{rao2021studying, chefer2021transformer}. As a quantitative evaluation of GeoT, the public benchmark for molecular property prediction MD17, QM9, and OC20 were used.

\subsection{The interpretative analysis of GeoAttention visualization from the chemical perspective}

Firstly, we provide a concise overview of $\text{RBF}_{embed}$ and AttnScale, which are the main components for understanding the mechanism of GeoAttention. The $\text{RBF}_{embed}$ consists of $k$ multiple Gaussian functions with different centers to embed an atom-atom distance as a $k$-dimensional vector. This vector $\{g(\mathrm{r})\}_k$ is then added to the sum of two atom-embedding vectors, $Z(a_i)+Z(a_j)$, to make $\boldsymbol{\Lambda} \in \mathbb{R}^{n \times k}$. Accordingly, the query atom vector $\boldsymbol{Q}$, the key atom vector $\boldsymbol{K}$, and $\boldsymbol{\Lambda}$ are multiplied together, to make $\mathrm{Attn}_G(\boldsymbol{Q}, \boldsymbol{K},\boldsymbol{\Lambda})$.  
Second, we introduced AttnScale to intensify the high frequency (HF) signal ${A}'_{HF}$ feature from $\mathrm{Attn}_G(\boldsymbol{Q}, \boldsymbol{K},\boldsymbol{\Lambda})$. The HF signal in a molecule can be viewed as the interactions between two atoms locating close to each other. The result is $\mathrm{Attn}'_G(\boldsymbol{Q}, \boldsymbol{K}, \boldsymbol{\Lambda})=\mathrm{Attn}_G(\boldsymbol{Q}, \boldsymbol{K}, \boldsymbol{\Lambda}) + {A}'_{HF}$. Finally, the formulas of GeoAttention of GeoT is as follows. We illustrated it in Fig. \ref{fig:diff_architecture}, and further details are described in Methods.

\begin{equation}
\mathrm{GeoAttention}'(\boldsymbol{Q}, \boldsymbol{K}, \boldsymbol{V}, \boldsymbol{\Lambda}) =  \mathrm{Attn}'_G(\boldsymbol{Q}, \boldsymbol{K}, \boldsymbol{\Lambda}) \cdot \boldsymbol{V}
\end{equation}

We selected six molecules having different molecular geometries (ring or linear) and conjugation patterns. For these molecules, we report each pattern of attention distributions produced by GeoT trained on two types of label $\epsilon_{LUMO}$ and $H$ of QM9 dataset, respectively. For the sake of brevity, we will refer to "GeoT trained on $\boldsymbol{\epsilon_{LUMO}}$\textbf{[}$\boldsymbol{H}$\textbf{]} in the qm9 dataset" as \textbf{$\text{GeoT}_{\textit{LUMO}}$[$\text{GeoT}_{\textit{H}}$]}.

\subsubsection{Comparative analysis on the pattern of attention weights by two target objectives: molecular orbital and enthalpy}

As a preliminary step, we introduce a brief overview of the chemical theory to provide context for our analysis. 
According to thermodynamic theory, $H$ is a sum of various energies of intramolecular interactions and chemical bondings, which is mainly determined by $\sigma$-bondings. On the other hand, in terms of molecular orbital (MO) theory, $\epsilon_{LUMO}$ strongly depends on the specific factor: the conjugated $\pi$-bondings included in conjugated molecules. In general, $\pi$-bondings are weaker than $\sigma$-bondings. The resonance energy from $\pi$-bonding also contributes to the overall $H$, but its energy scale is less significant than that of $\sigma$-bonding \cite{reusch1977introduction}.

The shape of electron density distributions associated with $\sigma$- and $\pi$-bondings are also different. The electron density of $\sigma$-bonding always localizes between the two atoms involved in the bonding. On the other hand, consecutive $\pi$-orbitals are conjugated to construct LUMO of a molecule. Accordingly, the electron density delocalizes (spread) over the molecular scaffold. More details are provided in Appendix \ref{Appendix.figures}.

Motivated by the contrasting characteristics between two chemical concepts, we visualized the attention maps produced by $\text{GeoT}_{\textit{LUMO}}$ and $\text{GeoT}_{\textit{H}}$ in Table \ref{tbl:lumoh}. Then we compared the observed patterns in terms of the theoretical expectations of two targets (delocalization for $\epsilon_{LUMO}$ and localization for $H$). 
To this end, we selected four molecules of different conjugation patterns: naphthalene, 1,2,3,4-tetrahydronaphthalene (tetralin), 1,3-dimethyl-2-(1,3-butadienyl)benzene, and $n$-decane. 

\textbf{Completely conjugated ring}. Naphthalene is a fully conjugated molecule composed of two fused hexagonal rings, represented in the first row \textbf{(a)}. In this case, we observed that the attention maps produced by $\text{GeoT}_{\textit{LUMO}}$ spread out all over the molecular scaffold alongside with the conjugated double bonds. In sharp contrast, the attention maps produced by $\text{GeoT}_{\textit{H}}$ are more localized around the C–H single bond in which the query atom is involved. 
These contrasting trends are consistently found across various types of molecules and query atoms. It indicates that the distribution of GeoAttention weight properly reflects the chemistry theory of MO, rather than depending on the individual query atoms. 

\textbf{Partially conjugated ring}. Tetralin has the same scaffold with naphthalene, but one of the hexagonal ring is not involved in the conjugation, represented in the second row \textbf{(b)}. The attention maps of tetralin highlights that conjugated scaffold is critical for the spread of attention weights, which is analogous to delocalization phenomenon in $\pi$-bondings. When the query atom is selected from the conjugated part of the molecule, the corresponding attention weights produced by $\text{GeoT}_{\textit{LUMO}}$ was mainly distributed only inside the conjugated part, which is similar to delocalization of molecular orbital. In contrast, when the query was selected from the non-conjugated part, the corresponding attention weights are localized around the query atom. Similarly, the attention maps from $\text{GeoT}_{\textit{H}}$ could not spread out from where the query atoms were selected. We emphasize the result because it is the strong evidence that \textbf{GeoT can differentiate aromatic and non-aromatic rings}, which have almost similar shape to each other.

\textbf{Conjugated compound}. 1,3-dimethyl-2-(1,3-butadienyl)benzene has an aromatic ring attached with conjugated butadiene and non-conjugated dimethyl, represented in the third row \textbf{(c)}. Similar to tetralin case, the corresponding attention weights spread out the conjugated region following the LUMO of a molecule if a conjugated atom is selected as the query. However, if a non-conjugated atom is selected, then the attention weights are localized and do not attend LUMO. It is another example that GeoT can understand the behavior of LUMO and distinguish conjugated atoms from a molecule.

\textbf{Non-conjugated linear compound}. For the last, we picked a decane as a contrasting example in the last row \textbf{(d)}. Decane is not conjugated, and accordingly there is no $\pi$-bonding. In this case, the distribution patterns from $\text{GeoT}_{\textit{LUMO}}$ and $\text{GeoT}_{\textit{H}}$ are similar to each other. It is also clear case that GeoT can understand that the close relation between conjugation and LUMO.

These results show that GeoT has an ability to consider “the significance of atomic pairs” at the appropriate region of a molecular scaffold in predicting the target molecular properties, as if GeoT understood the related chemical theory. 

\subsubsection{The effect of AttnScale on GeoAttention}

Next, we analyzed the effect of the modifications on the distribution of attention scores. Specifically, we assumed that the introduction of AttnScale can indeed improve the interpretability of GeoT by emphasizing more significantly related atoms to the target objective. We picked $H$ as the target objective because this distinguishing effect of AttnScale can be meaningful to interpret the attention weights of $\text{GeoT}_{\textit{H}}$, which showed more localized attention weights in Table \ref{tbl:lumoh}. We also compared the strengths of attention weights from the 3rd and 8th GeoAttention block, respectively. Note that we did not analyze the effect of $\text{RBF}_{emb}$ on the attention map, because the training loss of GeoT was not converged without $\text{RBF}_{emb}$ regardless of target type.

Table \ref{tbl:mod} shows two different conjugated molecule cases which were originated from $\text{GeoT}_{\textit{H}}$. We extracted the attention map from the 3rd and 8th GeoAttention block in both GeoT with AttnScale (\textbf{$\text{GeoT}_{\textit{H}}$}) and one without AttnScale (\textbf{$\text{GeoT}_{\textit{H}}\text{-base}$}), respectively. 

In both cases, the attention distribution is more concentrated to the query atom with $\text{GeoT}_{\textit{H}}$. On the contrary, the distribution is spread over whole molecule in the case of $\text{GeoT}_{\textit{H}}\text{-base}$. 
Considering two aspects that 1) the bonding energy is the most significant factor of determining $H$ of a molecule and 2) most of bonds exist between neighboring atoms, we conclude that the AttnScale can increase the attention scales between more closely located query and key atoms. This results also accord with the motivation of AttnScale \cite{wang2022anti}, which is the boosting of feature with HF signals, because it can help recognize the localized features from a query.

In addition, we found that the strength of AttnScale effects depends on the location of GeoAttention block in GeoT. Specifically, AttnScale effects are more salient in the 8th GeoAttention block rather than in the 3rd block, in both of molecules. This result also supports the conclusion of previous study \cite{wang2022anti} that AttnScale can prevent Transformer-based model from performance drop (over-smoothing problem), especially with deep architecture. The more cases are represented in Appendix \ref{Appendix.figures}.

\subsection{The case study: the prediction of energy changes as a function of a dihedral angle of biphenyl}

Conformational changes of a molecule structure alter geometric attributes including dihedral angles, while preserving the atomic connectivity graph.
Because GeoAttention can explicitly learn long-range inter-atomic relationships, we expected that GeoT shows superior performance for predicting energy profiles of conformational transitions.

\begin{figure}
\centering

    \begin{subfigure}{.96\textwidth}
        \centering
        \includegraphics[width=.45\linewidth]{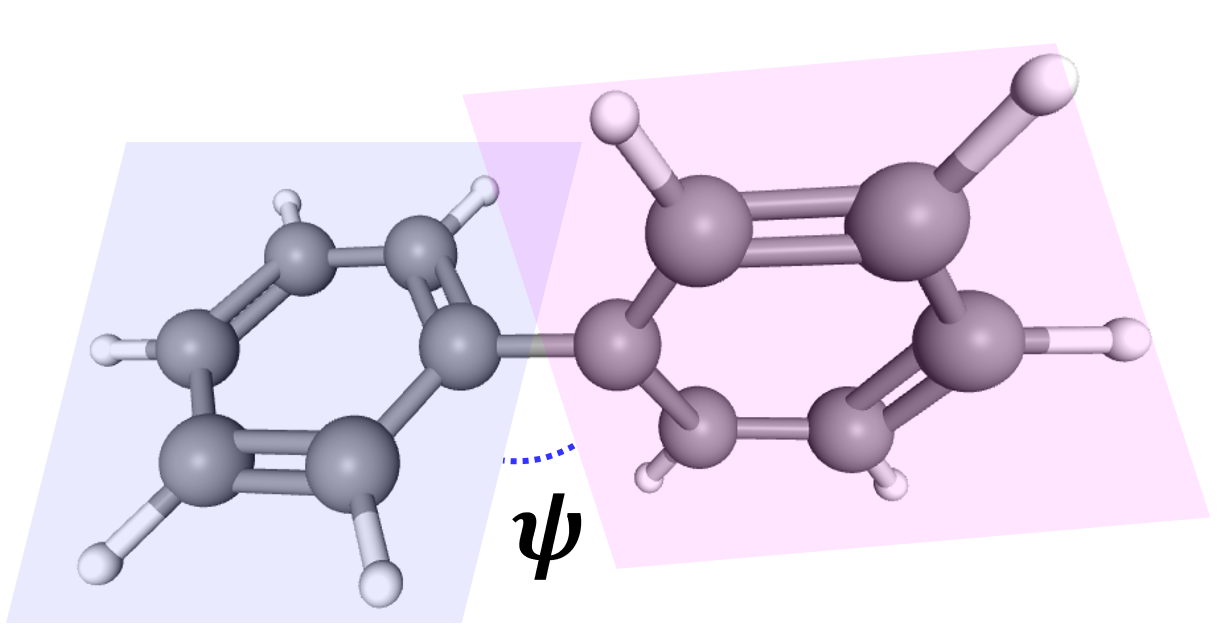}
        \caption{The conformation of a biphenyl molecule with its dihedral angle $\psi$}
        \label{fig:bip.3d}
    \end{subfigure}

    \begin{subfigure}{.48\textwidth}
        \centering
        \includegraphics[width=.95\linewidth]{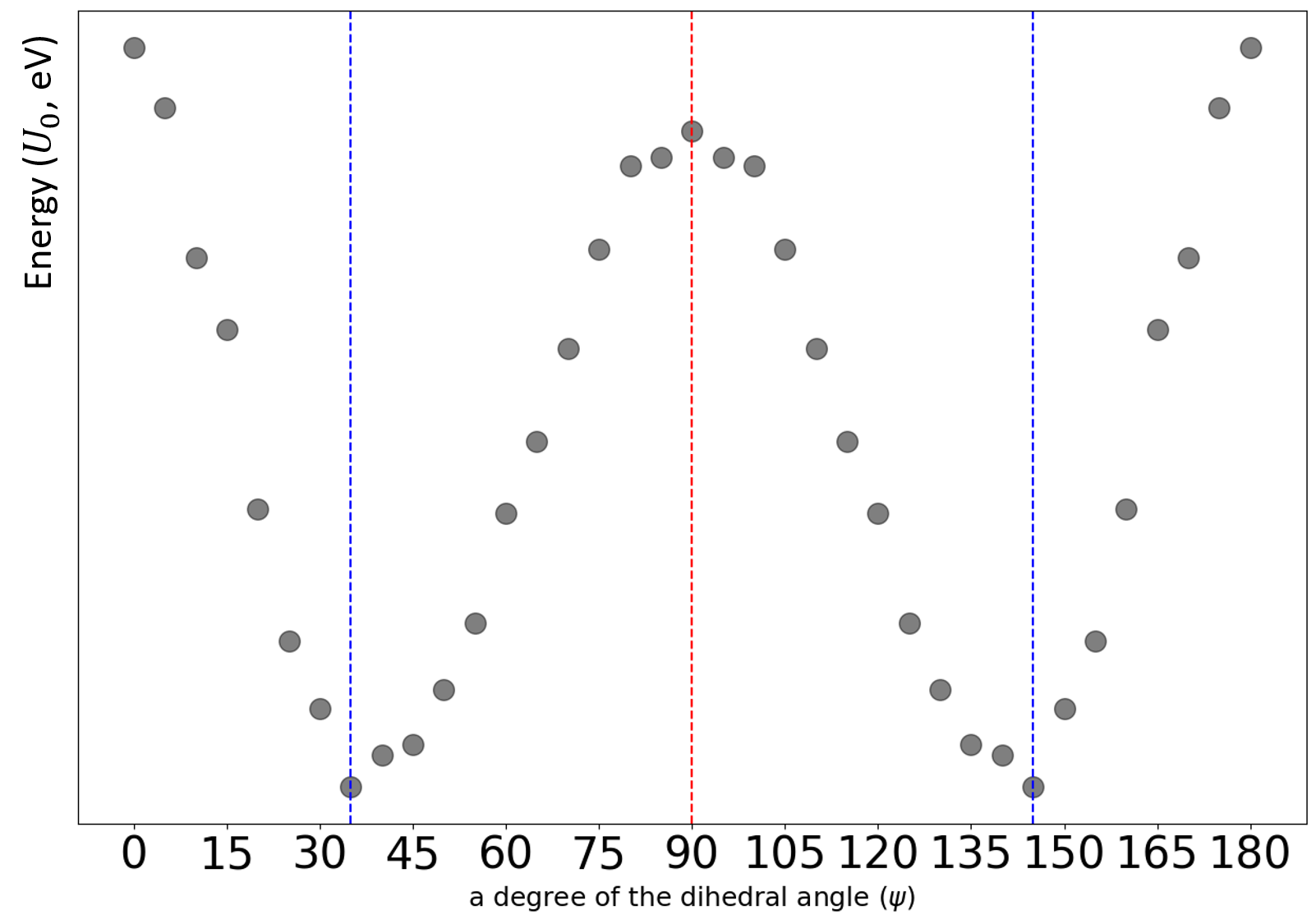}
        \caption{DFT (reference)}
        \label{fig:bip.dft}
    \end{subfigure}
    \hfill
    \begin{subfigure}{.48\textwidth}
        \centering
        \includegraphics[width=.95\linewidth]{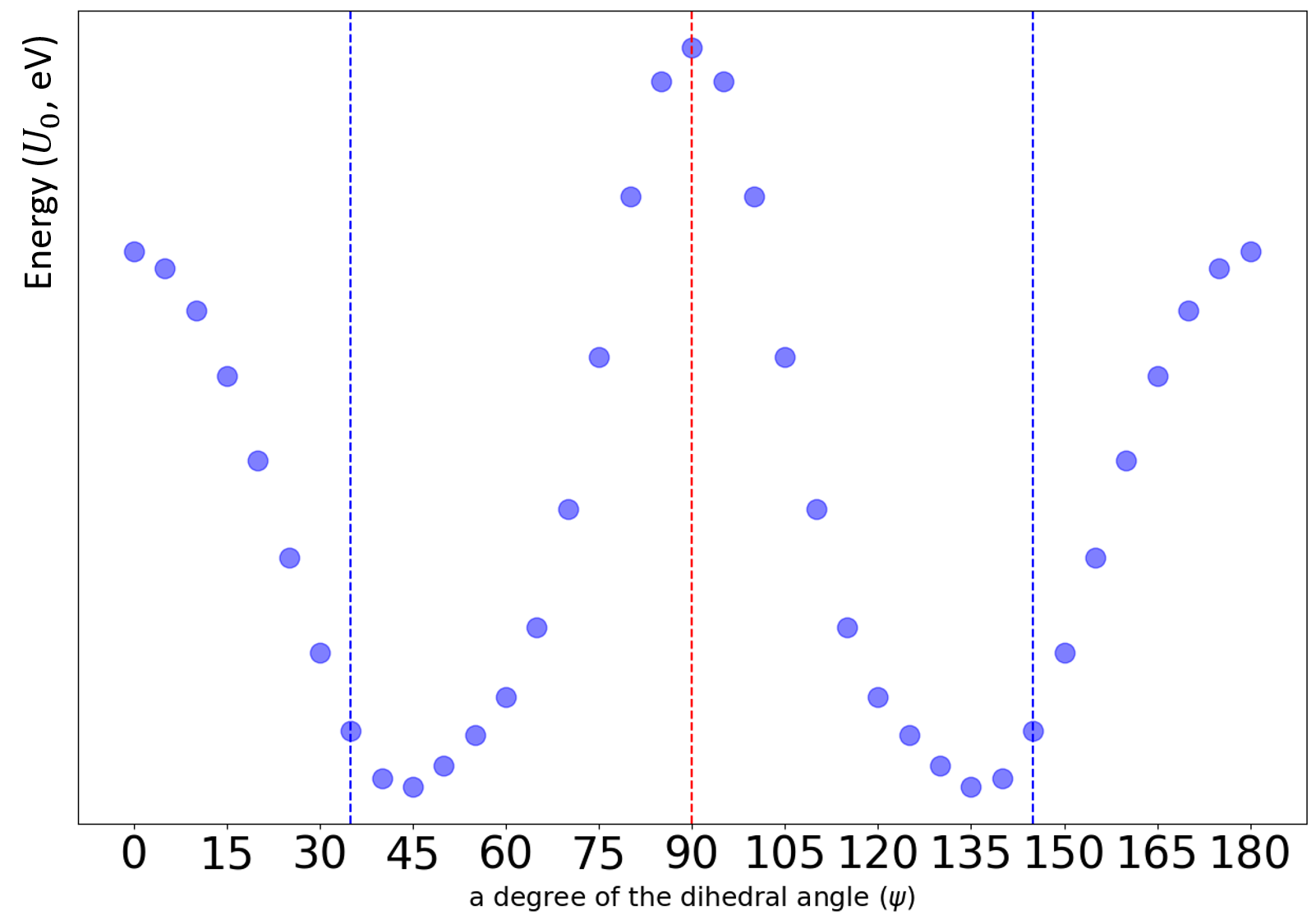}
        \caption{GeoT}
        \label{fig:bip.geot}
    \end{subfigure}

    \begin{subfigure}{.48\textwidth}
        \centering
        \includegraphics[width=.95\linewidth]{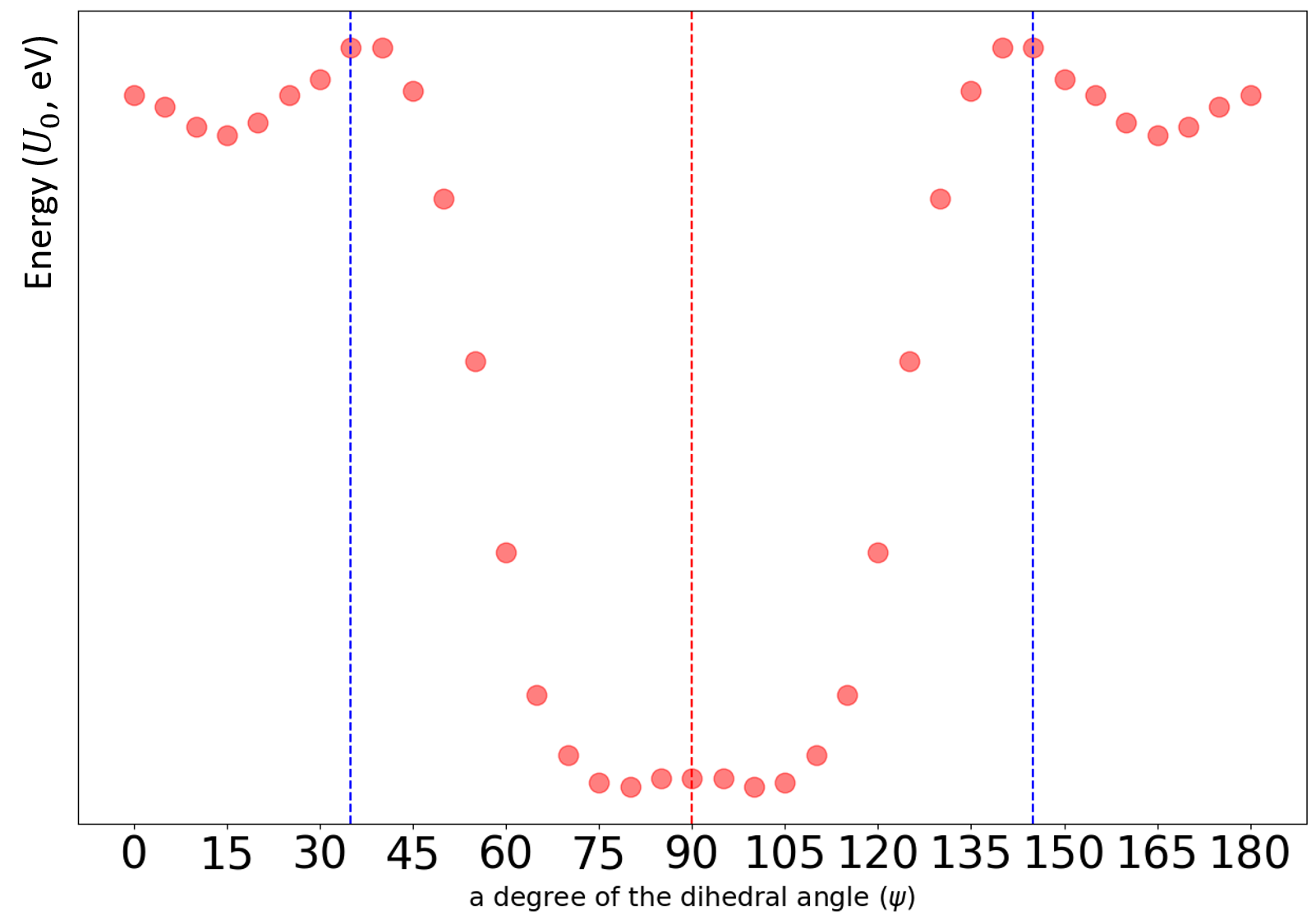}
        \caption{SchNet}
        \label{fig:bip.schnet}
    \end{subfigure}
    \hfill
    \begin{subfigure}{.48\textwidth}
        \centering
        \includegraphics[width=.95\linewidth]{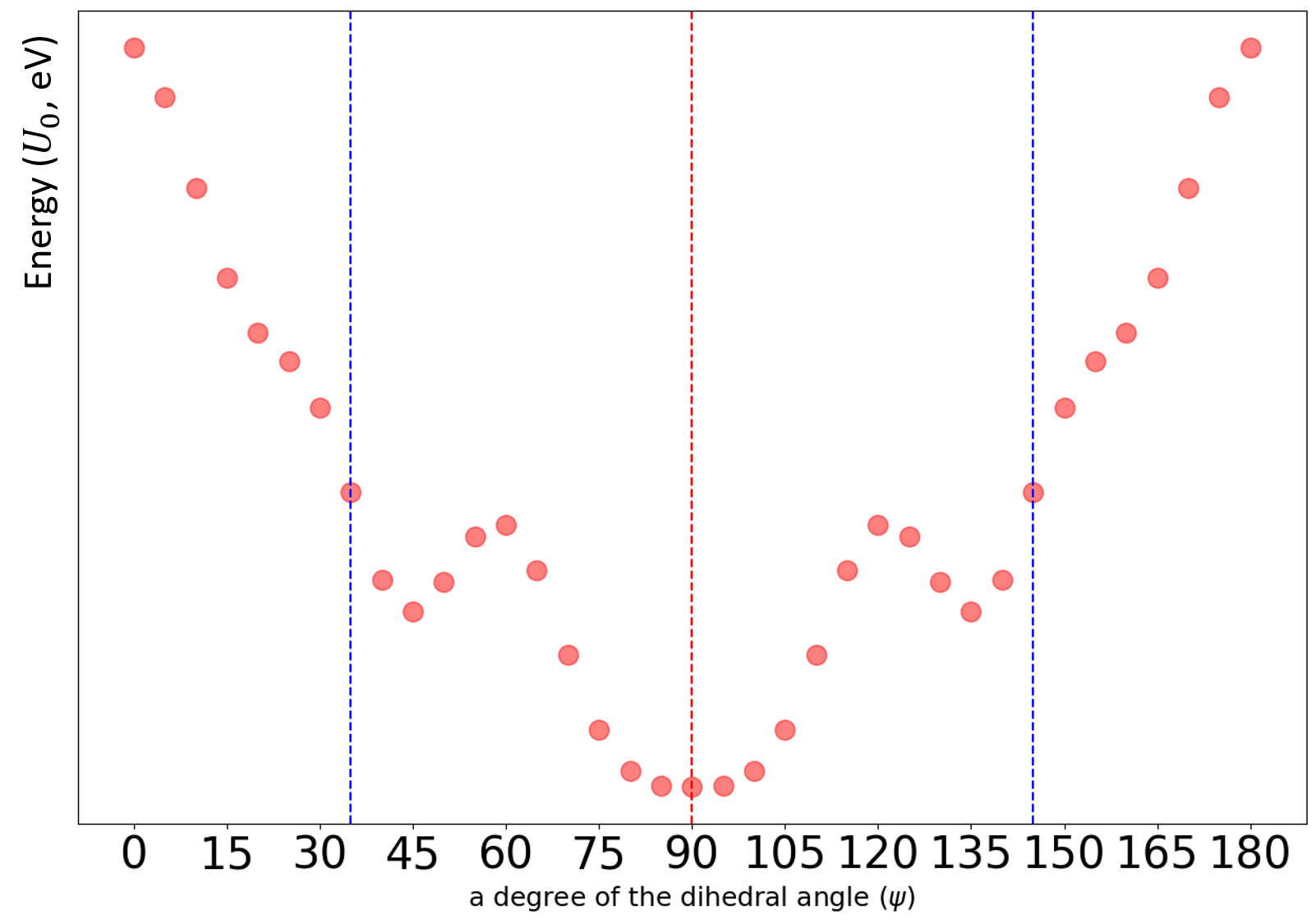}
        \caption{DimeNet}
        \label{fig:bip.dimenet}
    \end{subfigure}

    \caption{The plot of internal energy ($U_0$, eV) of a biphenyl molecule as a function of the dihedral angle ($0^\circ \leq \psi \leq 180^\circ\text{, } \Delta \psi = 5.0^\circ$) obtained by different methods. (a) An illustration of a biphenyl molecule, (b) computed by DFT simulation, (c)-(e), predicted by GeoT, SchNet, and DimeNet, respectively. The red and blue dashed lines presented on the each graph highlight the $\psi$-coordinate where DF-computed global maximum ($\psi=90^\circ$) and minimum (about $\psi=35^\circ$ and $145^\circ$) locate, as the ground truth.}
    \label{fig:biphenyl}
\end{figure}

We chose the biphenyl molecule as a test case: it is composed of two benzene rings connected by a single bond, and its total energy is fairly sensitive to the change of the dihedral angle ($\psi$) between two rings (in Fig. \ref{fig:biphenyl}). This molecule has been actively studied because its rotational characteristic of a dihedral angle significantly affects the physical and chemical properties of a molecule \cite{grein2002twist, johansson2008torsional, jain2017biphenyls}. In theory, when the two rings are coplanar to each other ($\psi = 0^\circ$ or $180^\circ$), the repulsion between hydrogen atoms accounts for the energy increase of biphenyl. When the two rings become perpendicular to each other ($\psi$ approaches toward $90^\circ$), conjugation over the two aromatic system would be broken and the total energy will be (locally) maximized. The output from DFT simulation clearly supports the above theory as the reference (Figure \ref{fig:bip.dft}, and we compared the predictive performance for this energy landscape by GeoT, SchNet, and DimeNet (Fig. \ref{fig:bip.geot}, \ref{fig:bip.schnet}, and \ref{fig:bip.dimenet}), respectively.

Obviously, GeoT reproduced overall shape of the energy profile with better accuracy than other methods do. Especially, GeoT successfully predicted the angle ($\psi = 90^\circ$) of which the energy is maximized. 
GeoT also predicts the minimum points ($\psi = 35^\circ$ and $145^\circ$) with better accuracy than other methods. Clearly, this task is a representative case that all carbon atoms of the molecule are involved, and thus it can be achieved only by considering long-range interatomic relationships.

\subsection{Model performance on public benchmarks for molecular property prediction}

Table \ref{tbl:mae_md17} shows MAE values on MD17 dataset, which comprises the energy prediction tasks of various conformations of each molecule. We compared the result of GeoT with three previous results sGDML \cite{chmiela2019sgdml}, DimeNet \cite{klicpera2020directional}, and GemNet-T \cite{gasteiger2021gemnet} in the left side of the table. 
In the right side of Table, the effects of using $\text{RBF}_k$ introduced in Eq. \ref{eq:rbffunc}, and other two model refinement strategies paralleled MLP and AttnScale introduced in Section \ref{model refinement} are shown. All strategies contribute to improve the model performance in six out of eight types of molecules, except for Aspirin and Malonaldehyde. 
GeoT outperformed in three molecules: Benzene, Ethanol, and Malonaldehyde, whereas GemNet-T achieved the best performance on the rest five molecules.

Table \ref{tbl:mae_qm9} composes MAE on QM9 datasets of GeoT and five previous studies for comparison. SchNet \cite{schutt2017schnet}, Cormorant \cite{anderson2019cormorant}, PhysNet \cite{unke2019physnet}, and DimeNet++ \cite{klicpera_dimenetpp_2020} are message passing-based methods, and GRAT \cite{yoo2020graph} is the only Transformer-based model. Note that Cormorant \cite{anderson2019cormorant} and DimeNet++ \cite{klicpera_dimenetpp_2020} use angles as well as distances between atoms, whereas other three models \cite{schutt2017schnet, unke2019physnet, yoo2020graph} only use distances between atoms for prediction tasks, as GeoT does. Our model outperformed those three previous models which uses distance information only. However, the performance of our model does not exceed that of DimeNet++ \cite{klicpera_dimenetpp_2020}. GeoT achieved comparable performances with DimeNet++ on six targets and outperformed in one case. 

Table \ref{tab:oc2010k} presents the performance results in OC20 IS2RE (10k) dataset, compared with those of five previous models \cite{schutt2017schnet, klicpera2020directional, xie2018crystal, klicpera_dimenetpp_2020} reported in \cite{ocp_dataset}. Six types of ablation studies were conducted to validate three types of model refinement strategies, same as above.
We should mention that SchNet and DimeNet used the periodic boundary conditions (PBC) to represent repeated structures of surfaces. The three types of model refinements were not effective for the 10k task. Table \ref{tab:oc20full} presents the performance results of various versions of GeoT and competing OC20 IS2RE tasks with a size of 10k and full datasets, respectively.

\subsection{Ablation study}
The ablation study of various RBF types is shown in Appendix \ref{Appendix.tables}. We performed ablation studies on simple linear and, Gaussian basis functions, which are used in SchNet \cite{schutt2017schnet}, and as well as radial bases. Bessel basis functions, which are used in DimeNet \cite{klicpera2020directional, klicpera_dimenetpp_2020}, correspond to the expressions given in Table \ref{basis functions}.
The results in Table \ref{tbl:mae_ablation} show that the Gaussian basis functions produced the best performance with GeoT.

\section{Discussion}
\label{Discussion}

We developed GeoT, a novel Transformer-based model for molecular property prediction based on molecular conformation. Each GeoT encoder block has a GeoAttention module, which learns the relationship between all pairs of atoms with their type information of a molecule graph. 
GeoAttention is the modified self-attention block which incorporates atom-atom distance information and enhances the high-frequency signal from heterogeneous key atoms. To the best of our knowledge, it is the first interpretative analysis of the attention pattern derived in molecular graphs with geometric features, from the chemical perspective.

GeoT has three advantages over the previous studies. First, GeoT can visualize the contributions of all atom-atom relationships for determining property of a given molecule. Surprisingly, the attention map obtained from trained GeoAttention shows clearly distinctive patterns depending on the type of targets. The attention pattern from $\text{GeoT}_{\textit{LUMO}}$ follows resonance structure of $\pi$-bonding regardless of molecular shape, whereas those from $\text{GeoT}_{\textit{H}}$ are highly correlated with the $\sigma$-bonding. 

From those observations, we conclude that the attention can differentiate the scale of various types of energies, and more attend to feature with larger contribution to target objective. Extensive study on various shaped-molecules with another target types are needed in the future study.

Second, we conducted a case study on the biphenyl molecule to evaluate the performance of Transformer in predicting the energy distribution across conformational changes, including the identification of the maximum and minimum energy geometries. It means that GeoT can be generalized to predict the molecular energy in sub-optimal state, as well as those in the optimal state. Notably, GeoT also achieved remarkable performance on three public benchmarks: MD17, QM9, and OC20. In particular, GeoT showed comparable performances with other MPNN-based models, which need additional computations for angle values. Indeed, GeoT outperformed the previous MPNN models and other Transformer-based models, without use of angle values.

Third, GeoT is more computationally efficient than the MPNNs that use angle values between three atoms. Unlike these models, our approach does not require a cutoff value and can learn long-range features without any restriction, providing an advantage in terms of computational efficiency over previous MPNN-based studies.

One limitation of GeoT is that the prediction performances on the public benchmarks are not consistently superior to those of the previous models, although GeoT has a lower computational cost than other models requiring angle computation and extra embedding spaces. Another point is that GeoT is capable of providing interpretability from a relative perspective with respect to an arbitrarily selected query atom, but not from a global perspective. This is due to the fact that GeoT is based on the Transformer architecture, which is designed to consider the relative relationships between components of a data.

To overcome those limitations of GeoT, we will seek to improve prediction performance on molecule graphs, and advanced representation strategies of attention weights for gaining deeper chemical insights.

\section{Methods}
\label{Methods}

In this section, we provide the preliminaries and descriptions of GeoT architecture, and the experimental details and training strategies.

\subsection{Preliminary: Transformer encoder}
\label{Model}

We provide the basics of original self-attention mechanism and Transformer encoder proposed in \cite{vaswani2017attention}. We do not describe decoders, because our methods based only on the encoder part of the original Transformer.

\textbf{Self-attention.} The self-attention mechanism allows the Transformer to focus on different elements of the input depending on the task. A set of input vectors $\{\boldsymbol{X}^l_i\}_{i \in \{1, ..., N\}}$ is split into three types of tensors: query $\boldsymbol{Q}$, key $\boldsymbol{K}$, and value $\boldsymbol{V}$, respectively. In practice, each matrix is obtained by the multiplication of individual weights $W_Q, W_K, \text{ and } W_V$ on the same input $X$, respectively.

\begin{equation} 
\label{eq:qkvmatrix}
\boldsymbol{Q} = W_{Q}\boldsymbol{X} ,\;\; \boldsymbol{K} = W_{K}\boldsymbol{X} ,\;\; \boldsymbol{V} = W_{V}\boldsymbol{X}
\end{equation}

After that, the dot-product is executed between $\boldsymbol{Q}$ and $\boldsymbol{K}$, and scaled by the vector dimension $d_{m}$.

\begin{equation} 
\label{eq:attention_calc_prev}
\mathrm{Attention}(\boldsymbol{Q}, \boldsymbol{K}, \boldsymbol{V}) = \mathrm{softmax}(\boldsymbol{QK}^T / \sqrt{d_{m}}) \cdot \boldsymbol{V}
\end{equation}

\textbf{Multi-head self-attention.} It is common to implement multiple individual self-attention blocks in Transformer-based architecture, because it can be advantageous for tasks depending on the complex interactions between different parts of the data, such as natural language processing.
Each matrix $\{\bm{Q}, \bm{K}, \bm{V}\}$ is split into $h$ matrices $\{\boldsymbol{Q}_i, \boldsymbol{K}_i, \boldsymbol{V}_i\}_{i \in \{1, ..., h\}}$ with dimension $d_{m} / h$, respectively. Self-attention is then applied to each $\{\boldsymbol{Q}_i, \boldsymbol{K}_i, \boldsymbol{V}_i\}_{i \in \{1, ..., h\}}$, followed by concatenation.

\begin{equation}
\label{eq:MSA}
\begin{split}
\boldsymbol{H}_i & = \mathrm{Attention}(\boldsymbol{Q}_i, \boldsymbol{K}_i, \boldsymbol{V}_i) \;\;\;\; i \in \{1, ..., h\} \\
\mathrm{MSA}(\boldsymbol{X}) & = \mathrm{Concat}(\boldsymbol{H}_1, ..., \boldsymbol{H}_h) \\
\end{split}
\end{equation}

\textbf{Transformer encoder layer.} The Transformer encoder layer is constructed by the layer normalization (LayerNorm) \cite{ba2016layer} and a feedforward ($\mathrm{FF}$) layer with skip connections. First, the output from $\mathrm{MSA}(\boldsymbol{X}^l)$ added to the original input $\mathbf{X}^l$ by the skip connection with a LayerNorm. After that, the last output $ \tilde{\mathbf{X}}^{l}$ is fed into $\mathrm{FF}$ layer, consisting of two linear transformations with an ELU \cite{clevert2015fast} activation. Finally, another LayerNorm is applied to the output from $\mathrm{FF}$ layer added by $\tilde{\mathbf{X}}^{l}$. This can be formulated as:

\begin{equation}
\label{eq:transformer_encoder_layer}
\begin{split}
\tilde{\mathbf{X}}^{l} & = \mathrm{LayerNorm}(\mathrm{MSA}(\boldsymbol{X}^l) + \boldsymbol{X}^l) \\
\boldsymbol{X}^{l+1} & = \mathrm{LayerNorm}(\mathrm{FF}(\tilde{\mathbf{X}}^{l}) + \tilde{\mathbf{X}}^{l})
\end{split}
\end{equation}

\subsection{Details of GeoT and GeoAttention}

We denote a molecule as a set of $N$ atoms $ \{a_i\}_{i \in \{1, \ldots, N\}}$ with their coordinates $\{\mathbf{r_i}\}\in \mathbb{R}^3$. We calculate the Euclidean distance $r_{ij} = \lVert \mathbf{r_{i}} - \mathbf{r_{j}} \rVert_2 $ between two atoms $a_i$ and $a_j$. Our model is a stack of $L$ layers, which transforms input $\boldsymbol{X}^l$ into output $\boldsymbol{X}^{l+1}$, especially the first layer input  $\boldsymbol{X}^0$ is given from the atom type embedding layer.

\subsubsection{Combining RBF into GeoAttention}

A single RBF $\mathbf{g}(r_{ij})$ used in GeoT is a Gaussian function which is defined by $\mathbf{g}(r_{ij})\colon \mathbb{R} \rightarrow  \mathbb{R}^{d_{m}}$, where an input $r_{ij}$ is the Euclidean distance between two atoms $a_i$ and $a_j$ and $d_{m}$ is the dimension of distance embedding. 
RBF is analogous to an embedding layer for a continuous-valued input set.

The $k$ multiple RBFs $\mathbf{g}_{k}(r)$ of the different centers of Gaussian distribution $\delta k$.
We selected multiple different Gaussian basis functions for constructing $\mathbf{g}_{k}(r)$, as proposed in SchNet \cite{schutt2017schnet}, where $\gamma=10$ and $\delta = 0.1 \textup{~\AA}$ are predefined constants.

\begin{equation}
\label{eq:shcnetbasis}
\mathbf{g}_{k}(r_{ij}) = \exp(-\gamma(||r_{ij} - \delta k||)^{2}) \;\;
\end{equation}

With $\mathbf{g}_{k}(r_{ij})$, we created the RBF embedding matrix $\boldsymbol{\Lambda}$ depending distance $r_{ij}$ with two feedforward layers $f_\theta$ as follows. We also embed atom types $a_i$ and $a_j$ into $f_\theta$, because atomic interaction depends on both the distance between two atoms and their atom types. To achieve this, we combined atom type embedding vector $Z(a_i)$  and $Z(a_j)$ of two atom $i$ and $j$ to $\mathbf{g}_{k}(r_{ij})$.
The $\mathbf{g}_{k}(r_{ij})$ is defined at the start of network, and the output $\boldsymbol{\Lambda}$ is individually provided in each GeoAttention block (detailed in Figure \ref{fig:general_architecture}). The matrix form of constructing $\Lambda$ is defined as below, where $\mathrm{\bf{r_{ij}}}$ and $\mathrm{\bf{a}}$ are the matrix representations of given atom indices $\{r_{i,j}\}_{i,j=1,...,N}$ and $a_{i=1,...,N}$, respectively.

\begin{equation}
\label{eq:rbffunc}
\boldsymbol{\Lambda} = f_{\theta}(\mathbf{g}_{k}(\mathrm{\bf{r_{ij}}}) \oplus  (Z(\mathrm{\bf{a_i}}) + Z(\mathrm{\bf{a_j}}))) 
\end{equation}

 Accordingly, the base form of GeoAttention is defined as below.
 
 \begin{equation}
 \begin{split}
 \mathrm{Attn_G}(\boldsymbol{Q}, \boldsymbol{K}, \boldsymbol{\Lambda}) &= (\boldsymbol{QK}^T\boldsymbol{\Lambda})/\sqrt{d_{m}} \\
 \mathrm{GeoAttention}(\boldsymbol{Q}, \boldsymbol{K}, \boldsymbol{V}, \boldsymbol{\Lambda}) &=  \mathrm{Attn}_G(\boldsymbol{Q}, \boldsymbol{K}, \boldsymbol{\Lambda}) \cdot \boldsymbol{V}
 \end{split}
 \end{equation}

\subsubsection{Enhancing the HF term in GeoAttention}
\label{model refinement}
AttnScale \cite{wang2022anti} is a type of modification for self-attention, which was originally developed to amplify HF signal of self-attention weights in deep Transformer architecture. The original paper \cite{wang2022anti} consider the direct-current (DC) component of self-attention $\boldsymbol{A}$ as $\bm{A}_{DC} = \frac{1}{N}\bm{1}\bm{1}^T$, and the other components are defined as HF terms $\boldsymbol{A}_{HF}$. As described above, AttnScale rescales HF term by $(1 + w_a)$, preventing from the over-smoothing effect which preserves only indistinguishable DC components in feature maps especially in deep Transformer architecture. 

Following from the original paper, we enhanced $\boldsymbol{A}_{HF}$ of GeoAttention, which can be naturally correspond to interatomic relations between closely located atom pairs in our tasks. The enhanced HF signal $\boldsymbol{A}'_{HF}=(1+w_a)\boldsymbol{A}_{HF}$, where $w_a \geq 0$. Consequently, the detailed 
 definition of $\mathrm{Attn}'_G$ is defined as below. 

\begin{equation}
\label{eq:AttnScale2}
\begin{split}
\mathrm{Attn}'_G(\boldsymbol{Q}, \boldsymbol{K}, \boldsymbol{\Lambda}) &= \boldsymbol{A}_{DC} + \boldsymbol{A}'_{HF} \\
 & =   \boldsymbol{A}_{DC} + (1+w_a)\boldsymbol{A}_{HF} \\
 & = \frac{1}{N} \sum_{k} \boldsymbol{A}_{k} + (1+w_a)\boldsymbol{A}_{HF} 
\end{split}
\end{equation}

\subsubsection{Construction of GeoAttention as multi-heads}

GeoAttention is defined by the multiplication of $\boldsymbol{V}$ by the above term $\mathrm{Attn'_G}(\boldsymbol{Q}, \boldsymbol{K}, \boldsymbol{\Lambda})$. We implemented GeoAttention as $H$ heads to achieve multi-head GeoAttention (denoted as $\{\mathrm{GeoAttention}\}_{H}$) by splitting each $\boldsymbol{Q},\;\boldsymbol{K},\;\boldsymbol{V}$ into $H$ components, applying individual GeoAttention to each component set, and concatenating them, where $h \in \{1, ..., H\}$ means a head index. 

\begin{equation}
\label{eq:GeoT.MSA}
\begin{split}
\boldsymbol{G}_{h} & = \mathrm{GeoAttention}(\boldsymbol{Q}_h, \boldsymbol{K}_h, \boldsymbol{V}_h, \boldsymbol{\Lambda}) \\
\{\mathrm{GeoAttention}\}_{H}& = \mathrm{Concat}(\boldsymbol{G}_{1}, ..., \boldsymbol{G}_{H}) \\
\end{split}
\end{equation}

For the sake of brevity, we omitted $H$ term: all GeoAttention described in this paper mean the multi-head term $\{\mathrm{GeoAttention}\}_{H}$.

\subsubsection{Parallelization in GeoT encoder}
For ensuring stable convergence of GeoT during training, we further modified the structure of GeoT encoder. Inspired by the previous work analyzing self-attention \cite{park2022vision}, we implemented only one layer normalization which takes (multi-head) GeoAttention, FF layers, and skip connection simultaneously. A block of GeoT encoder is formulated as below, where $l \in \{1,...,L\}$ means an encoder index.
\begin{equation}
\label{eq:TransformerencoderlayerParallel}
\begin{split}
\boldsymbol{X}^{l+1} = \mathrm{LayerNorm}( \mathrm{GeoAttention}(\boldsymbol{Q}, \boldsymbol{K}, \boldsymbol{V}, \boldsymbol{\Lambda}) + \mathrm{FF}(\boldsymbol{X}^{l}) + \boldsymbol{X}^{l})
\end{split}
\end{equation}

\subsubsection{Readout layer of GeoT for prediction output}

We introduced a sum-pooling layer at the final step, to obtain scalar-valued molecular property by aggregating atom features. This is a common strategy in MPNNs \cite{schutt2017schnet, klicpera2020directional, gasteiger2021gemnet}, which is named as the "readout" layer.

\begin{equation}
\label{eq:outputscalar}
\hat{y} = W_{\mathrm{L}}\boldsymbol{X}^L + b \;\;
\end{equation}

\subsection{Dataset}

We describe three benchmarks: MD17 \cite{chmiela2017machine, chmiela2019sgdml, chmiela2018towards}, QM9 \cite{ruddigkeit2012enumeration, ramakrishnan2014quantum}, and OC20 \cite{ocp_dataset}. The original MD17 dataset \cite{chmiela2017machine} comprises 10 small molecules. For each molecule, the dataset contains energy and forces for different geometries with more than 10k structures, based on DFT calculations. Instead of using original dataset, we used a dataset comprising 1,000 training samples \cite{chmiela2018towards} with more precise calculations than the original MD17 dataset. The task of the datasets is to predict the forces and energy of a given conformation from other conformations of specific molecules.

QM9 comprises 134k small organic molecules consisting of carbon, hydrogen, nitrogen, oxygen, and fluorine in an equilibrium state. All molecular conformations and their corresponding properties were created through computational simulations based on DFT calculation.
The datasets contains stable three-dimensional coordinates of each molecule and their 12 scalar quantum chemical properties, including geometric, energetic, electronic, and thermodynamic properties of given molecular structure. 
The task of the datasets is to predict the properties from molecular conformation.
As the datasets contain comprehensive chemicals with high consistence \cite{ramakrishnan2014quantum}, many molecular property prediction tasks were evaluated on the datasets.

Open Catalyst 2020 (OC20) \cite{ocp_dataset} is an open-source for learning catalysis dynamical properties with chemical configurations. Similar to the benchmarks mentioned above, OC20 data comprises a three-dimensional configurations with atomic numbers of each adsorbate onto surfaces in the initial and DFT relaxation states. Nevertheless, OC20 is a more advanced task as the relaxed state between adsorbates and surfaces must predict the initial state thereof. OC20 includes three different types of tasks: from structure to predict energy and forces (S2EF), from the initial structure to predict the relaxed structure (IS2RS), and from the initial structure to predict the relaxed energy (IS2RE). Each task is subdivided into several tasks according to the dataset size. We focused on the IS2RE tasks on 10k and full-sized datasets, which require relatively less computations.

\subsection{Visualization details}
\label{Methods:visualization}

We extracted $\Vert \mathrm{Attn}'_G(\boldsymbol{Q}, \boldsymbol{K}, \boldsymbol{\Lambda})\Vert_2$ of several molecules from two types of GeoT trained on QM9: $\text{GeoT}_{\textit{LUMO}}$ and $\text{GeoT}_{\textit{H}}$. Subsequently, an atom $i$ was chosen for each molecule as a query, and its GeoAttention values were realized using the other atoms $\{j\}\setminus i$. Query atoms are marked with blue asterisks, and attention norms of key atom $j$ from the query atom $i$ are colored with red shades. The shades are gradually colored by interpolations for visibility and easy comparison with MO. The blue circle around the query atom has a radius of 5\AA.

\subsection{A case study of biphenyl molecule}

GAMESS, which is the open source for computational chemistry analysis \cite{perri2014web, barca2020recent} was used to produce DFT simulation of the internal energy of biphenyl molecule. The 6-31G$^*$ was used as the basis set and DFT functional was B3LYP. We used the pretrained SchNet and DimeNet provided by Pytorch geometry \cite{Fey/Lenssen/2019} version 2.2, as the comparative study.

\subsection{Training}
In this study, the depth of GeoT was set to ($L=8,\;16$) to perform QM9 prediction, whereas it was four ($L=4$) each in the case of the MD17 and OC20 dataset.
We used 300 ($n_{\mathrm{basis}}=300$) different Gaussian functions with $\mu_{k} = 0.1 \cdot k$ to embed atom-atom distances. 
The dimension of GeoT was defined as (256, 512, 1024) with (4, 16, 64) multi-head GeoAttention ($H=4,\;16,\;64$).
The Swish \cite{ramachandran2017searching} and ELU \cite{clevert2015fast} activation function were used for the RBF and feed forward layers, respectively.

For QM9 training, we removed 3k molecules which were previously reported to have unstable conformation, following from the previous works \cite{schutt2017schnet, unke2019physnet, klicpera_dimenetpp_2020, anderson2019cormorant, yoo2020graph}.
The mean absolute error (MAE) was used to perform evaluations according to the guidelines, and each label was trained individually.
Adam \cite{kingma2014adam} was used as the optimizer with MAE loss and the batch size was 32 in all experiments.
The learning rate was set to 0.0002 at initial step, and decreased by 0.95 for every 200k steps with linear warmup 3,000 steps. We applied the early stopping method by evaluating every 10k step of training.
The maximum number of training epochs was up to 300.

To calculate forces, we assumed the predicted energy as a function of positions $E(\{ \mathbf{r}_i \})$. Based on the relation between force and potential energy of atoms, we calculated forces by differentiating energies with atom positions following from the previous works \cite{chmiela2019sgdml, schutt2017schnet}.
\begin{equation}
\label{eq:forcerelation}
F_{j} = \frac{\partial}{\partial \mathbf{r}_j}E(\{\mathbf{r}_i\})
\end{equation}
In the MD17 dataset, both the energy and forces of molecular conformations are provided. To utilize both features with Eq. \ref{eq:forcerelation}, we trained our model with the modified loss function with additional force terms as given by Eq. \ref{eq:compositeloss} \cite{chmiela2019sgdml, schutt2017schnet}. 
\begin{equation}
\label{eq:compositeloss}
L(E,\hat{E})=\lvert E-\hat{E} \rvert +c\times\sum_{j=i}^{n} \lvert F_j-\frac{\partial}{\partial\mathbf{r}_j}\hat{E}(\{\mathbf{r}_i\}) \rvert
\end{equation}

where $c$ is the weight coefficient for the loss on forces. We set $c$ as 1000 in our experiments.

For training the OC20 dataset, the trained model was evaluated on the in-distribution subset for validation, following the official guideline \cite{ocp_dataset}. Moreover, the periodic boundary condition (PBC) trick was not implemented here, whereas all the other methods implemented it. PBC is an approximation technique for analysis in large systems with small repeated patterns. With the PBC method, multiple replicas of a small unique pattern are arranged to represent a regular system. Consequently, broad regular surface molecules can be represented with a much smaller-sized unique cell. Thus, it significantly lowers the computations and ensures consistent representations of different atoms on the same surface. 

\subsection{Data availability}
We utilized the public open web sources (http://quantum-machine.org/, http://www.sgdml.org/, and https://opencatalystproject.org/) to obtain three public benchmark datasets QM9, MD17, and OC20, respectively. We provide the generated input coordinates of biphenyl.

\subsection{Code availability}
The code used for this study is available on GitHub https://github.com/oleneyl/geometry-aware-transformer.

\section{Acknowledgements}
This study was supported by Institute of Information \& communications Technology Planning \& Evaluation (IITP) grant (No.2021-0-01343, Artificial Intelligence Graduate School Program in Seoul National University; and No.2021-0-02068), and also supported by the National Research Foundation of Korea (NRF) funded by the Ministry of Education (2022R1A6A3A01087603, 2022R1A3B1077720, 2022M3C1A3081366). All supports were funded by the Korea government (MSIT).

\newcommand{\addpic}{\includegraphics[width=8em]{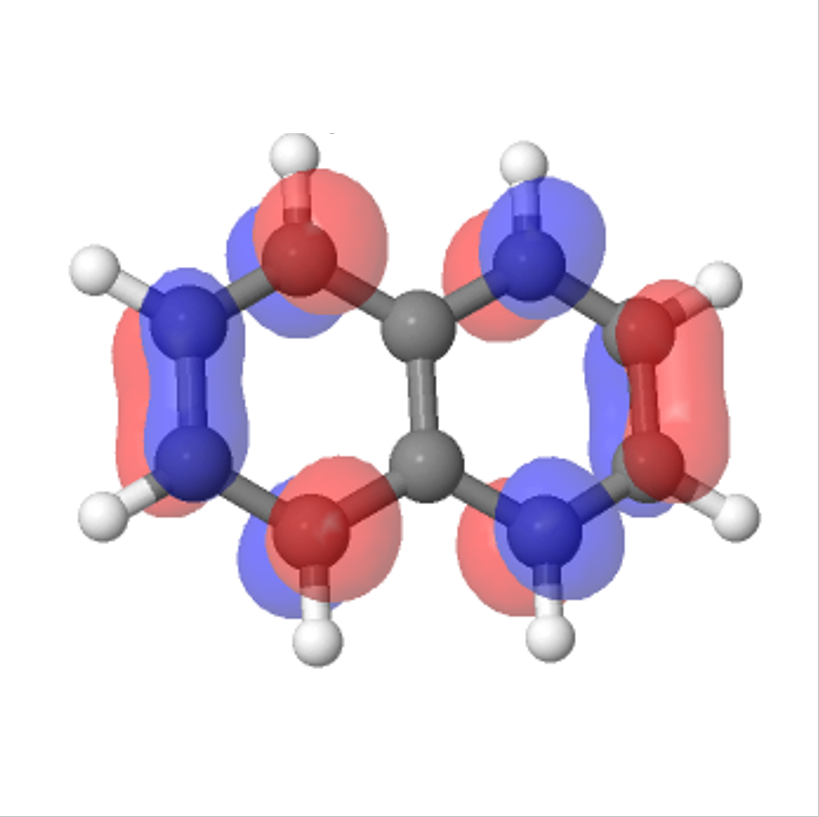}}
\newcolumntype{B}{>{\centering\arraybackslash}m{6em}}
\newcolumntype{C}{>{\centering\arraybackslash}m{12em}}
\begin{table}[!htb]\sffamily
\caption{The comparison between attention distributions of $\text{GeoT}_{\textit{LUMO}}$ and $\text{GeoT}_{\textit{H}}$ in the qm9 dataset, respectively. 
In each molecule, two individual query atoms are selected (marked by a blue asterisk). The attention weights are described by color gradients: more [\textcolor{red}{red}/yellow] shades represent [\textcolor{red}{stronger}/weaker] attention weights.} 
\label{tbl:lumoh}
\begin{tabular}{lB*2{C}@{}}
\toprule
Molecule & LUMO from DFT simulation (reference) & the attention distribution produced by $\text{GeoT}_{\textit{LUMO}}$ & the attention distribution produced by $\text{GeoT}_{\textit{H}}$ \\ 
\midrule
(a) & \includegraphics[width=.95\linewidth]
        {naphthalene_lumo2_square.png} & \includegraphics[width=.95\linewidth]
        {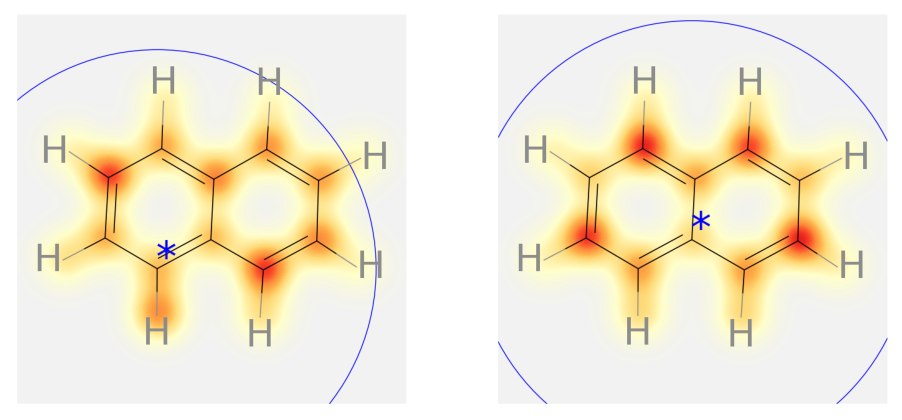} & \includegraphics[width=.95\linewidth]
        {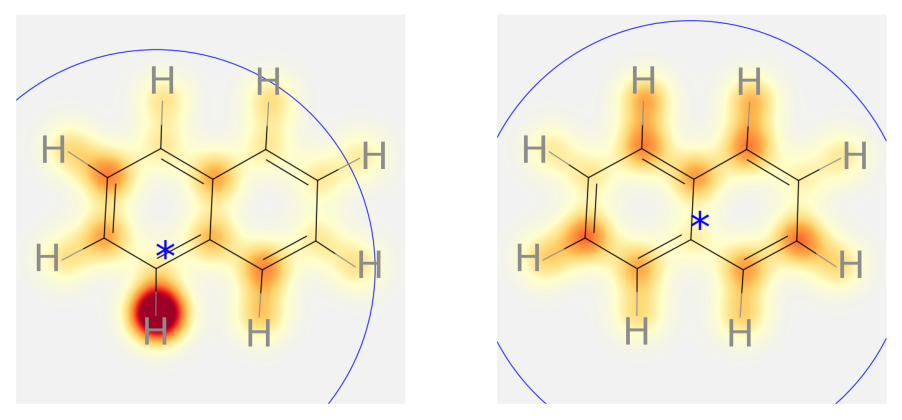} \\ 
(b) & \includegraphics[width=.95\linewidth]
        {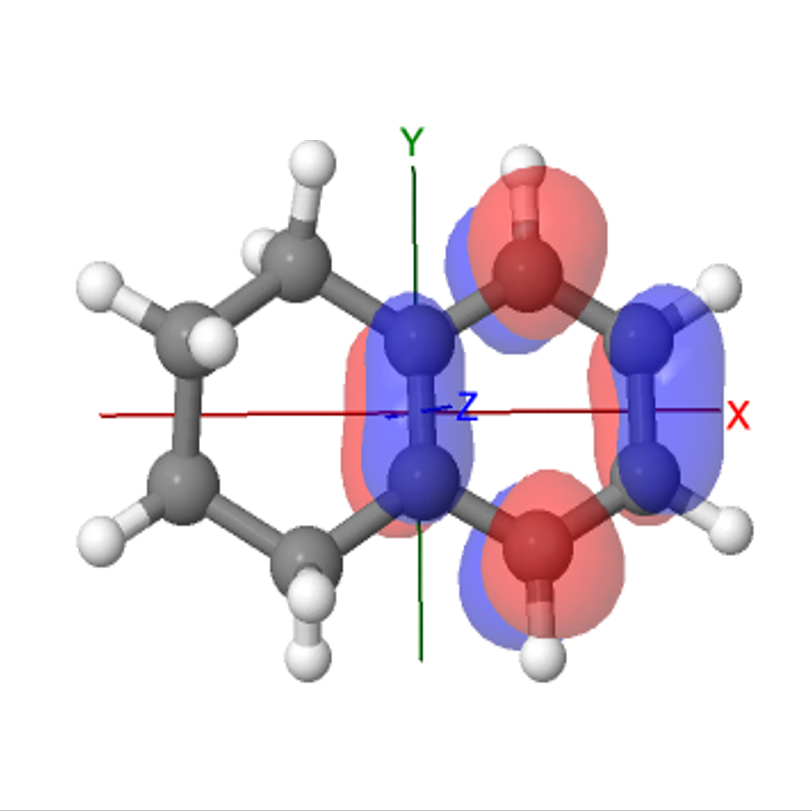} & \includegraphics[width=.95\linewidth]
        {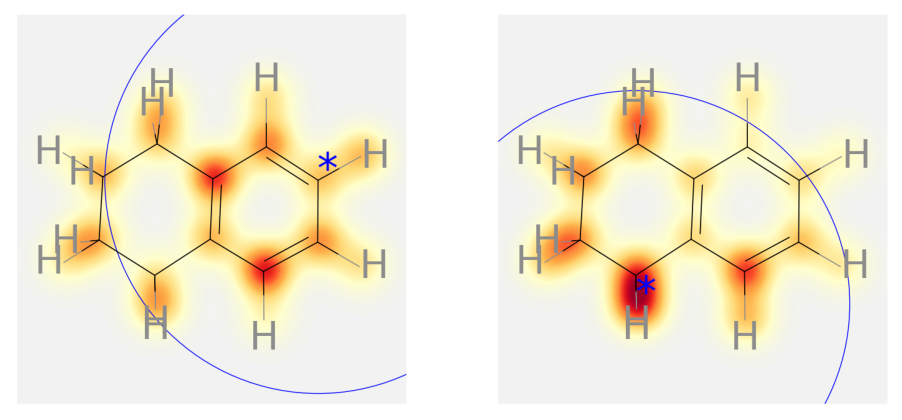} & \includegraphics[width=.95\linewidth]
        {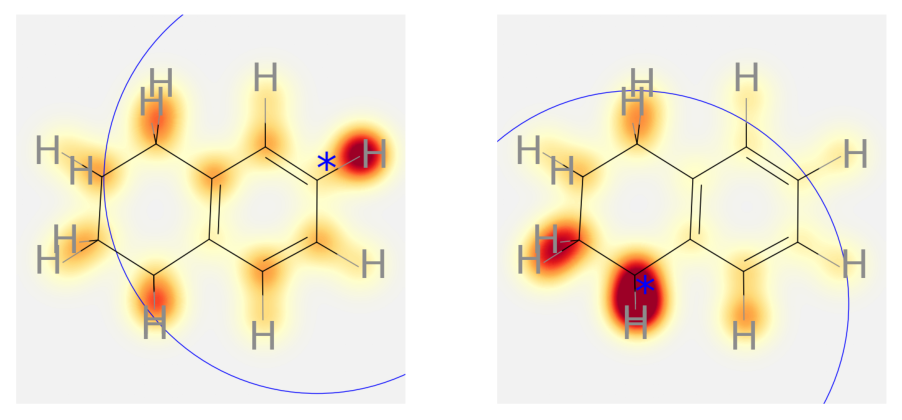} \\  
(c) & \includegraphics[width=.95\linewidth]
        {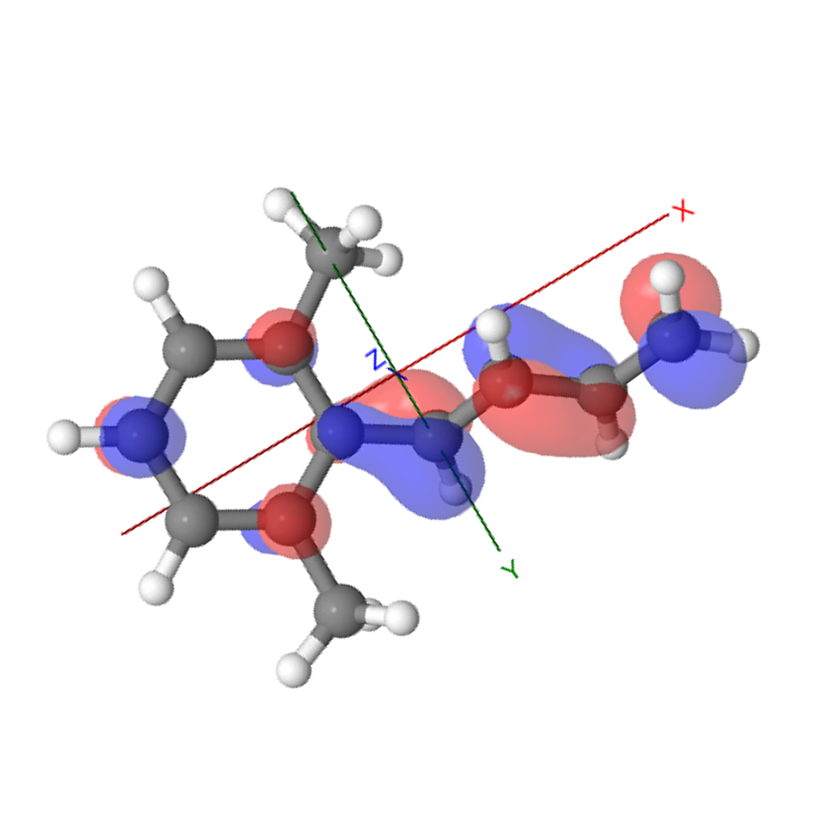} & \includegraphics[width=.95\linewidth]
        {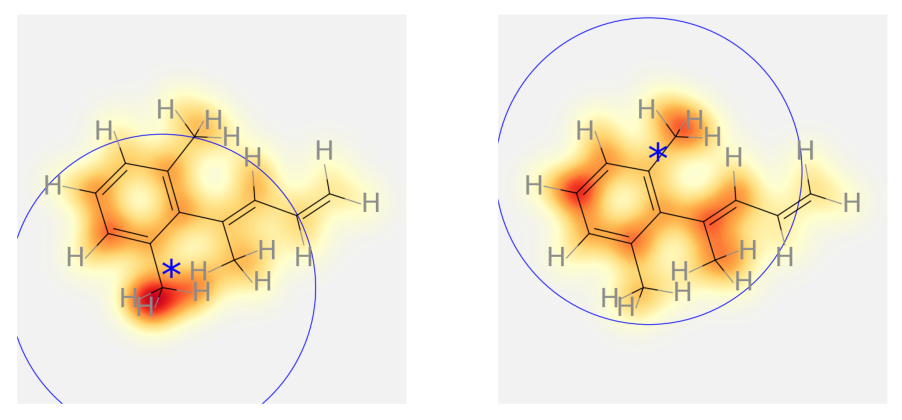} & \includegraphics[width=.95\linewidth]
        {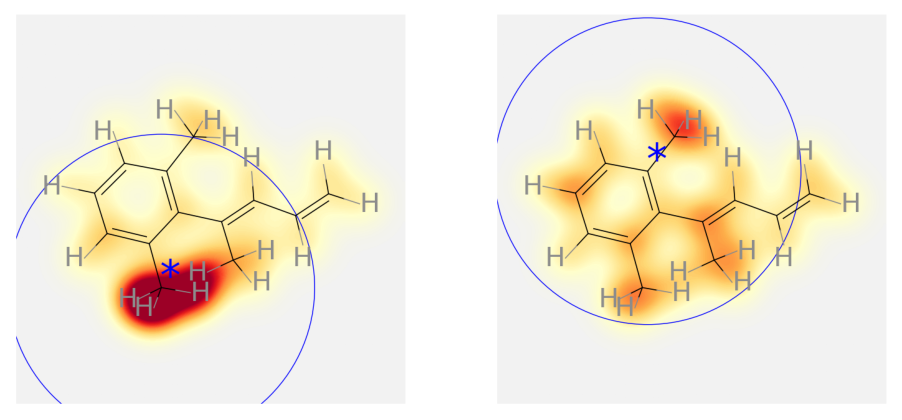} \\  
(d) & \includegraphics[width=.95\linewidth]
        {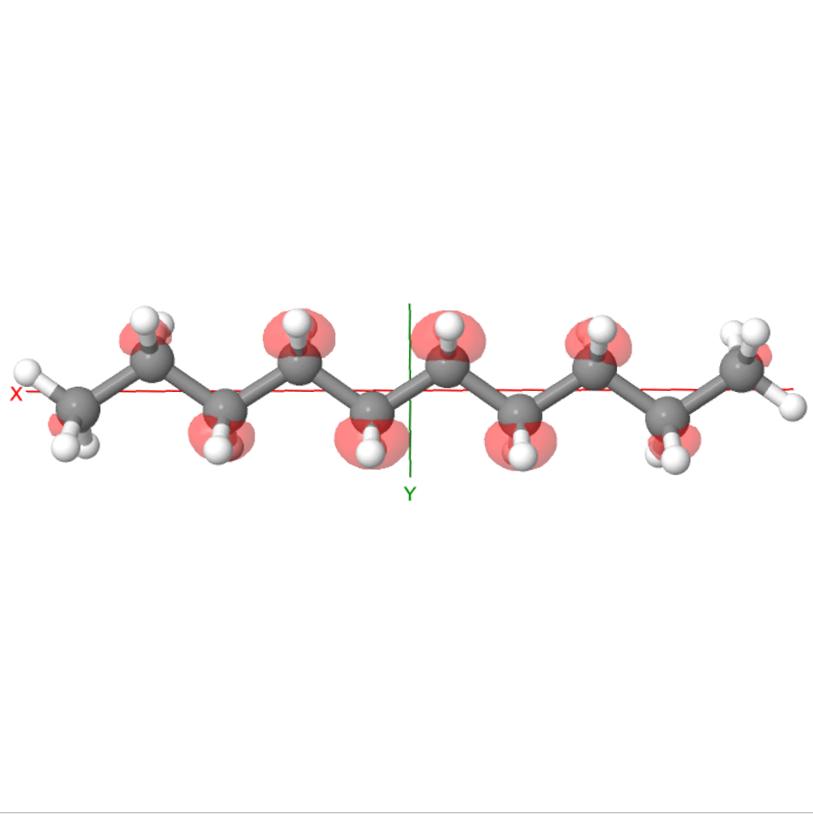} & \includegraphics[width=.95\linewidth]
        {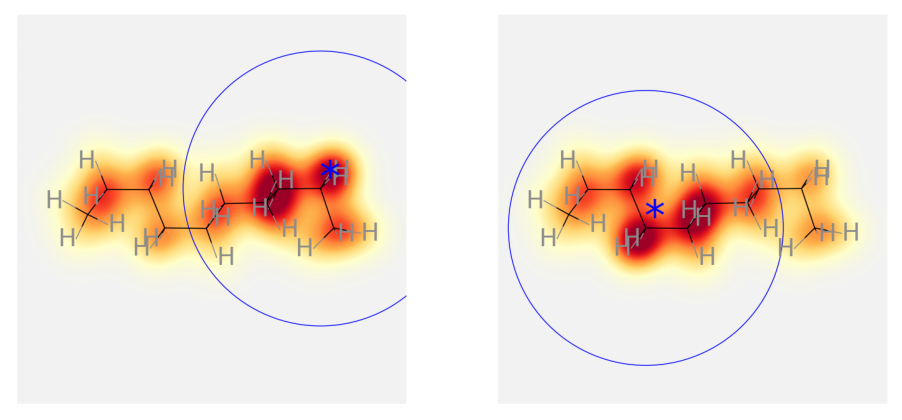} & \includegraphics[width=.95\linewidth]
        {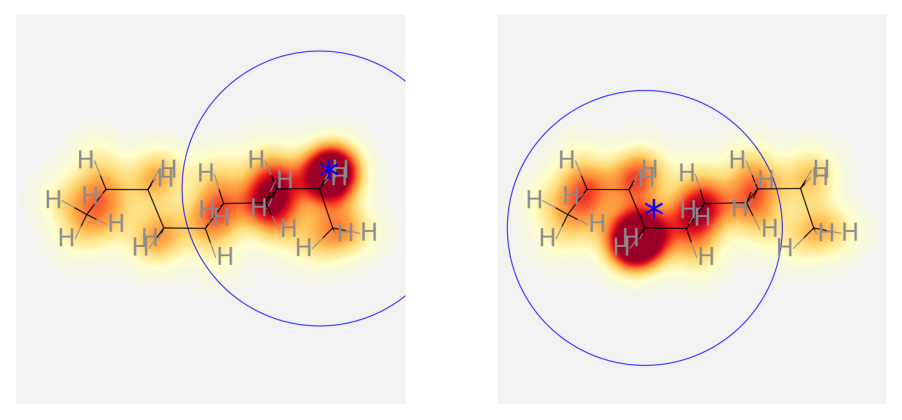} \\  
\bottomrule 
\end{tabular}
  \begin{tablenotes}
    \footnotesize
    \item The represented molecules are \textbf{(a)} naphthalene, \textbf{(b)} tetralin, \textbf{(c)} 1,3-dimethyl-2-(1,3-butadienyl)benzene, and \textbf{(d)} decane, respectively.
  \end{tablenotes}
\end{table}

\newcolumntype{D}{>{\centering\arraybackslash}m{7em}}
\newcolumntype{E}{>{\centering\arraybackslash}m{10em}}
\begin{table}[!htb]\sffamily
\caption{The comparative analysis on the effect of AttnScale of GeoAttention.  
Overall, the AttnScale make the attention weights be more concentrated around the query. The scale of AttnScale effect is larger in the 8th GeoAttention, rather than in the 3rd one. }
\centering
\label{tbl:mod}
\begin{tabular}{l*3{E}@{}}
\toprule
Molecule & the attention distribution produced by $\text{GeoT}_{\textit{H}}$ & the attention distribution produced by $\text{GeoT}_{\textit{H}}\text{-base}$ & The difference of $\text{GeoT}_{\textit{H}}-\text{GeoT}_{\textit{H}}\text{-base}$, (\textcolor{blue}{+}:blue, \textcolor{red}{-}:red) \\ 
\midrule
(a) & \includegraphics[width=.95\linewidth]
        {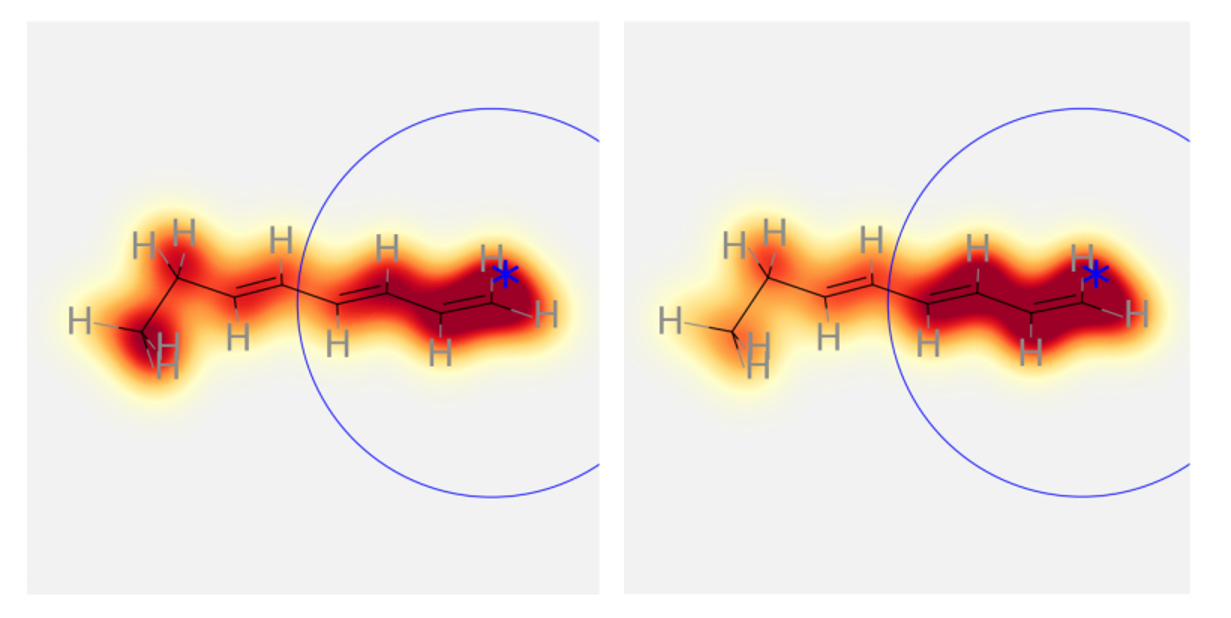} & \includegraphics[width=.95\linewidth]
        {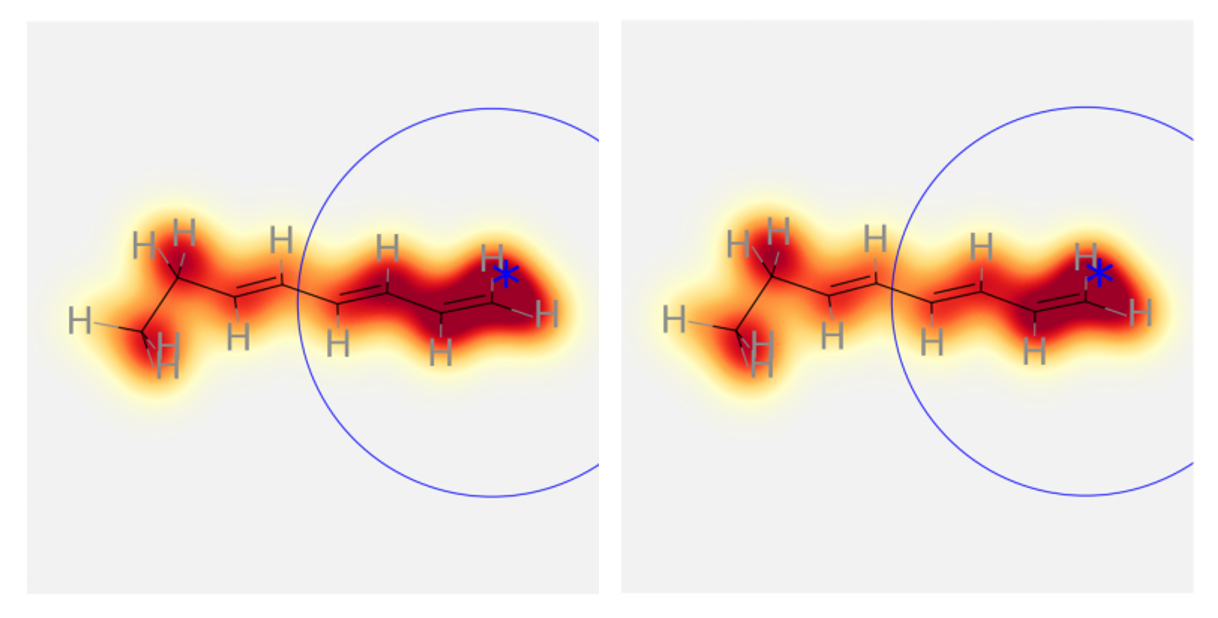} & \includegraphics[width=.95\linewidth]
        {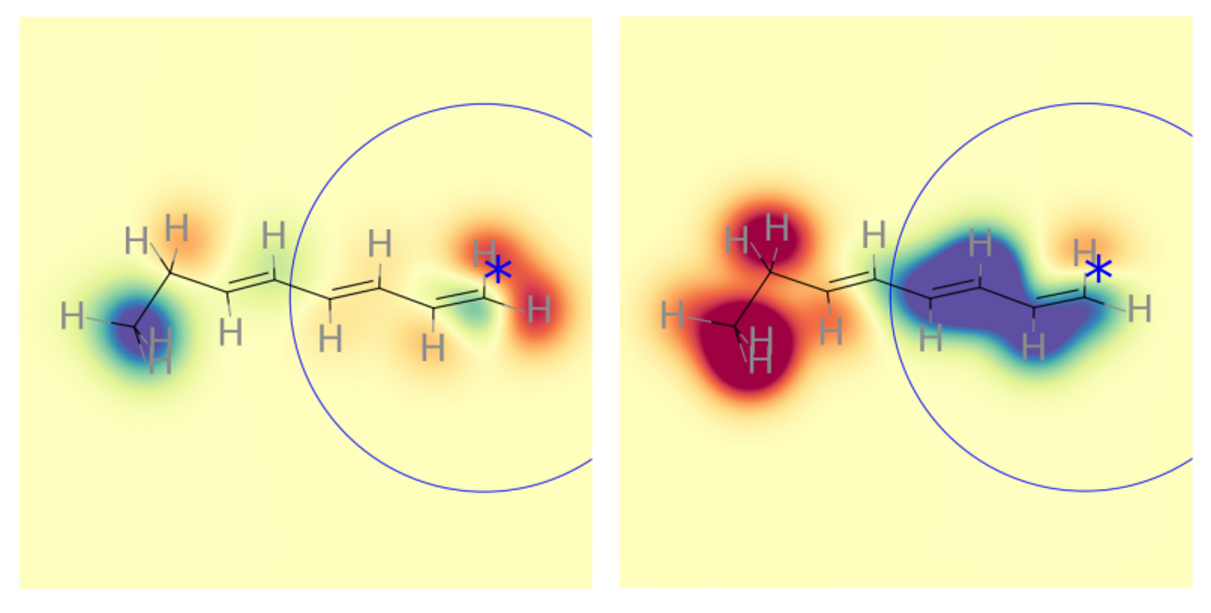} \\ 
(b) & \includegraphics[width=.95\linewidth]
        {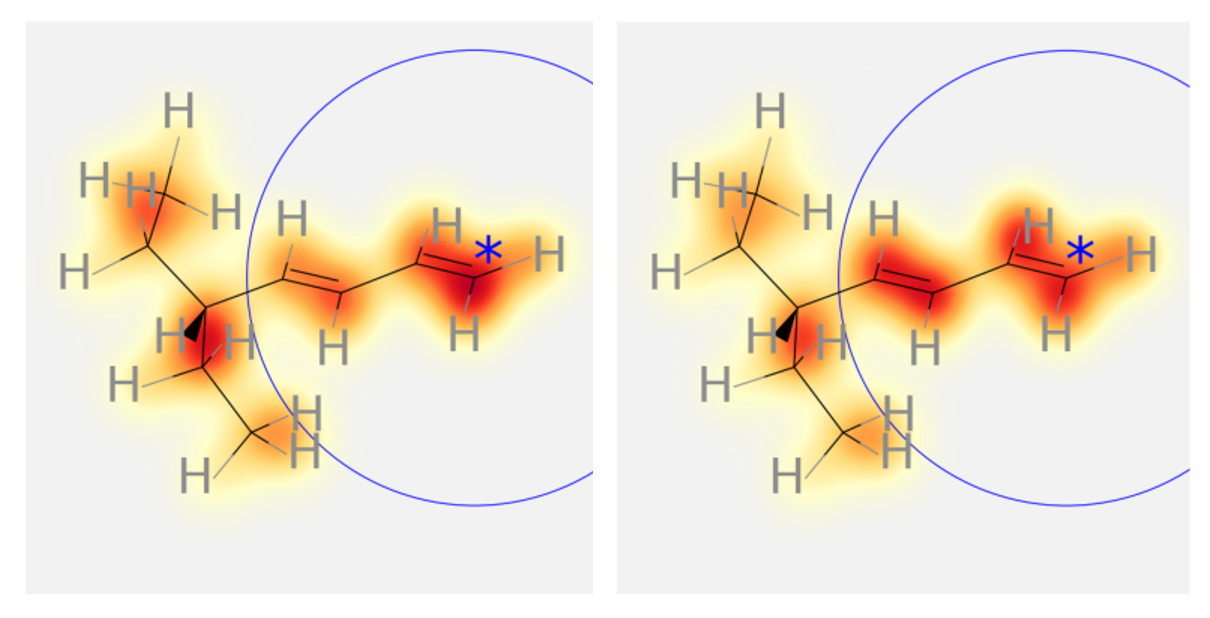} & \includegraphics[width=.95\linewidth]
        {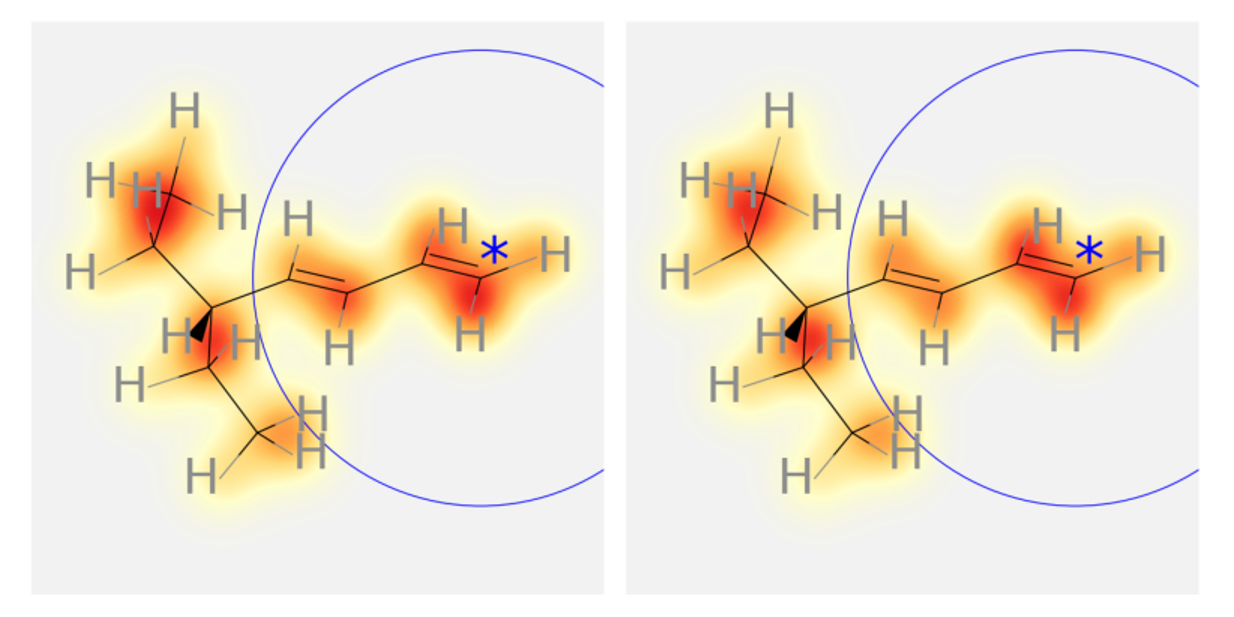} & \includegraphics[width=.95\linewidth]
        {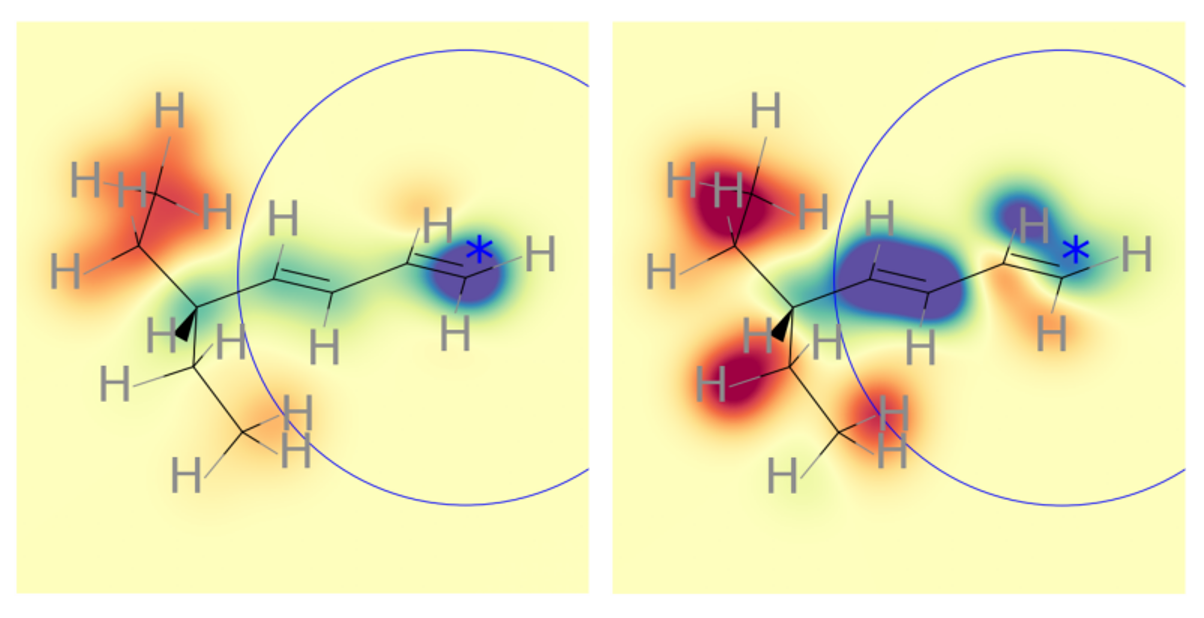} \\ 

\bottomrule 
\end{tabular}
  \begin{tablenotes}
    \footnotesize
    \item The represented molecules are \textbf{(a)} 1,3,5-octatriene, and \textbf{(b)} 7,7-dimethyl-1,3,5-octatriene, respectively.
  \end{tablenotes}
\end{table}

\begin{sidewaystable}
  \caption{Mean absolute errors (MAE) of the force prediction of MD17}
  \centering
  \label{tbl:mae_md17}
  \begin{adjustbox}{scale=0.90,center}

  \begin{tabular}{lccc|cccccc}
    \toprule
    \makecell{Molecule \\ Dependence on angles} & \makecell{sGDML \\ Yes } & \makecell{DimeNet \\ Yes } & \makecell{GemNet-T \\ Yes } & GeoT & GeoT+A & GeoT+B & GeoT+C & GeoT+B+C & GeoT+A+B+C
 \\
    \midrule
    Aspirin & 0.6803 & 0.4981 & 0.2191 & 0.7081 & 0.7147 & 0.7657 & 0.606 & 0.741 & 0.721 \\
    Benzene[9] & - & 0.1868  & 0.1453 & - & 0.10879 & 0.109 & 0.1132 & 0.1206 & 0.0978 \\
    Ethanol & 0.3298 & 0.2306 & 0.0853 & 0.09494 & 0.093508 & 0.08967 & 0.0952 & 0.07741 & 0.08599 \\
    Malonaldehyde & 0.4105 & 0.3828 & 0.1545 & 0.13985 & - & 0.1408 & 0.14215 & 0.1476 & 0.151 \\
    Naphthalene & 0.1107 & 0.2145  & 0.0553 & 0.2198 & 0.2264 & 0.1947 & 0.2321 & 0.1949 & 0.1924 \\
    Salicylic acid & 0.2790 & 0.3736  & 0.1268 & 0.2869 & 0.289 & 0.2849 & 0.2758 & 0.2755 & 0.2667 \\
    Toluene & 0.1407 & 0.2168  & 0.0600 & 0.1541 & 0.1538 & 0.1307 & 0.1642 & 0.1403 & 0.147 \\
    Uracil & 0.2398 & 0.3021 & 0.0969 & 0.1471 & 0.1499 & 0.1341 & 0.1631 &  0.13097 & 0.1285 \\
    \bottomrule
  \end{tabular}
  \end{adjustbox}
  \begin{tablenotes}
    \small
    \item `A' ,`B' and `C' mean the use of $\text{RBF}_k$, the parallelization of GeoT encoder and AttnScale, respectively.
    \item The measure is kcal/mol/A.
  \end{tablenotes}
\end{sidewaystable}

\begin{appendices}

\section{Additional visualization of GeoAttention}
\label{Appendix.figures}

\begin{figure}
\centering

    \begin{subfigure}{.45\textwidth}
        \centering
        \includegraphics[width=.95\linewidth]{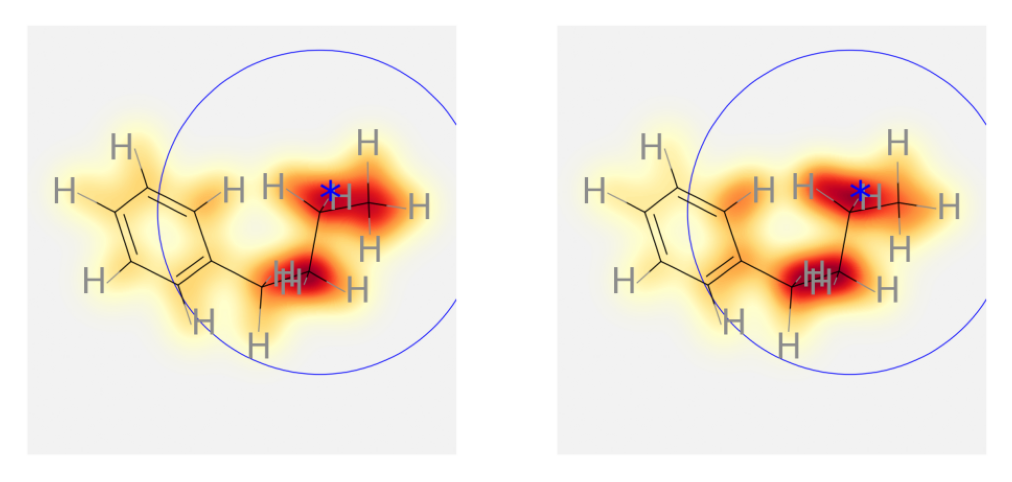}
        \label{fig:hlumo.butylester.lumo}
    \end{subfigure}
    \hfill
    \begin{subfigure}{.45\textwidth}
        \centering
        \includegraphics[width=.95\linewidth]{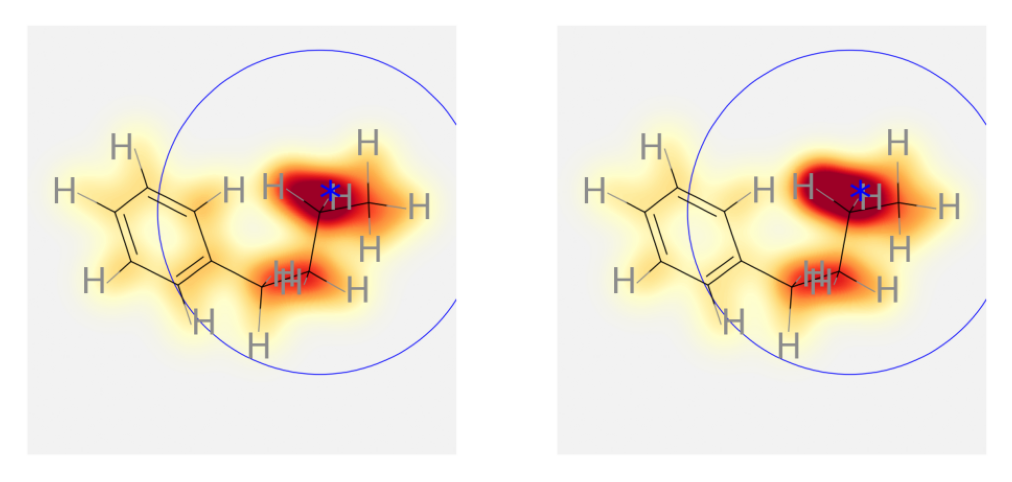}
        \label{fig:hlumo.butylester.h}
    \end{subfigure}

    \begin{subfigure}{.45\textwidth}
        \centering
        \includegraphics[width=.95\linewidth]{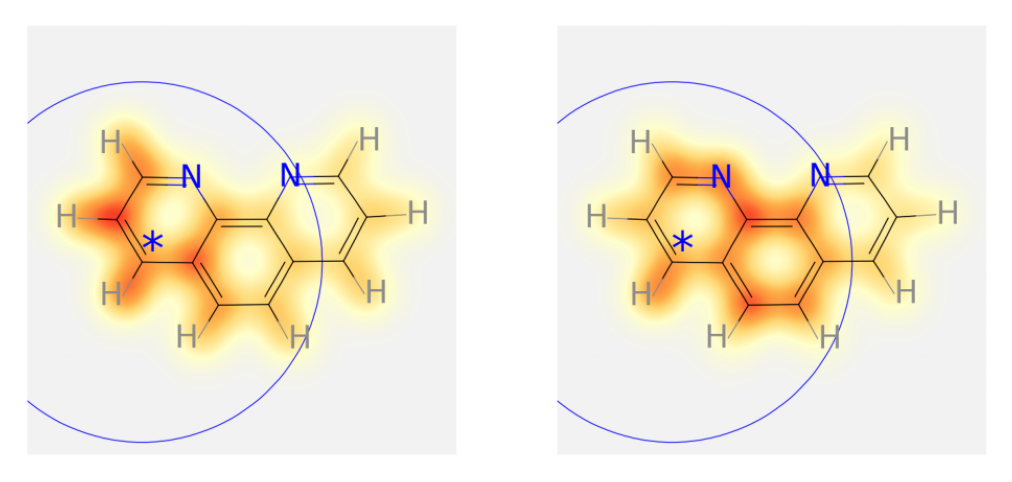}
        \label{fig:hlumo.phenanthroline.lumo}
    \end{subfigure}
    \hfill
    \begin{subfigure}{.45\textwidth}
        \centering
        \includegraphics[width=.95\linewidth]{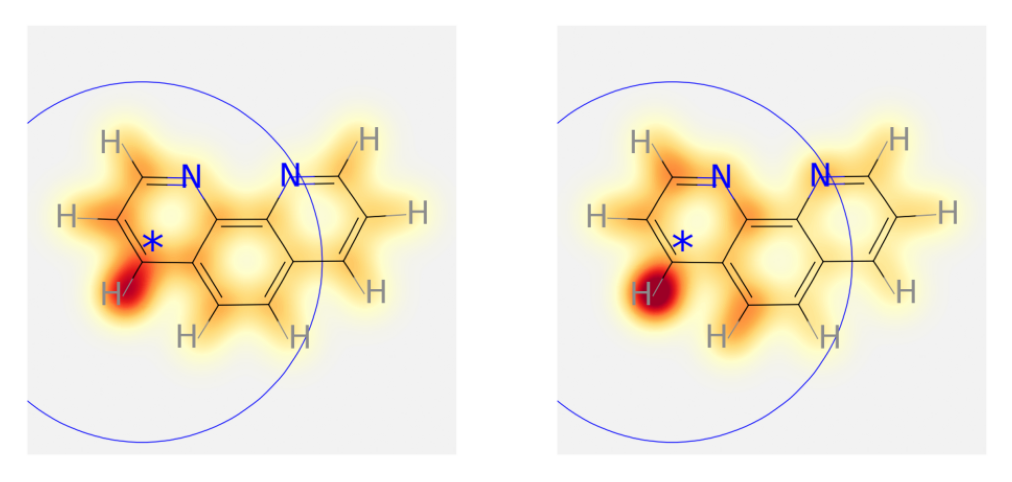}
        \label{fig:hlumo.phenanthroline.h}
    \end{subfigure}

    \begin{subfigure}{.45\textwidth}
        \centering
        \includegraphics[width=.95\linewidth]{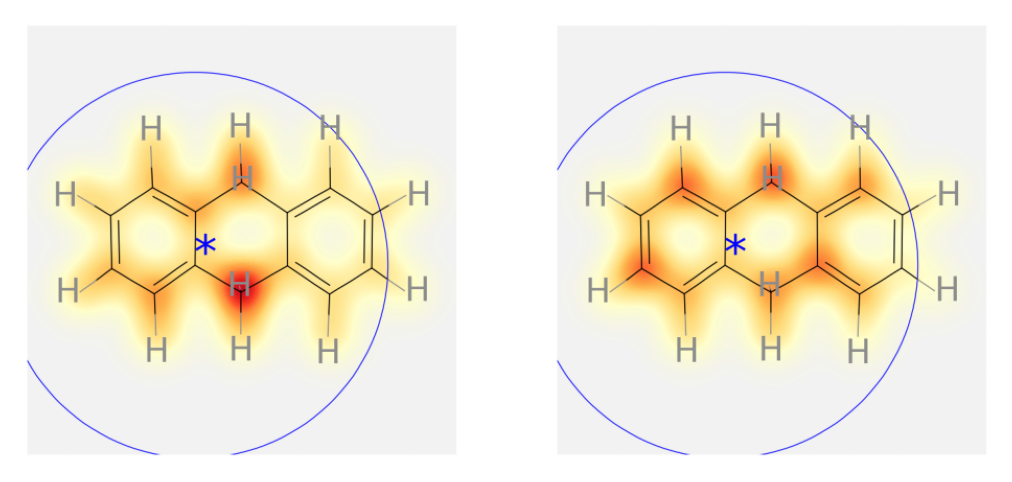}
        \label{fig:hlumo.dihydro.lumo}
    \end{subfigure}
    \hfill
    \begin{subfigure}{.45\textwidth}
        \centering
        \includegraphics[width=.95\linewidth]{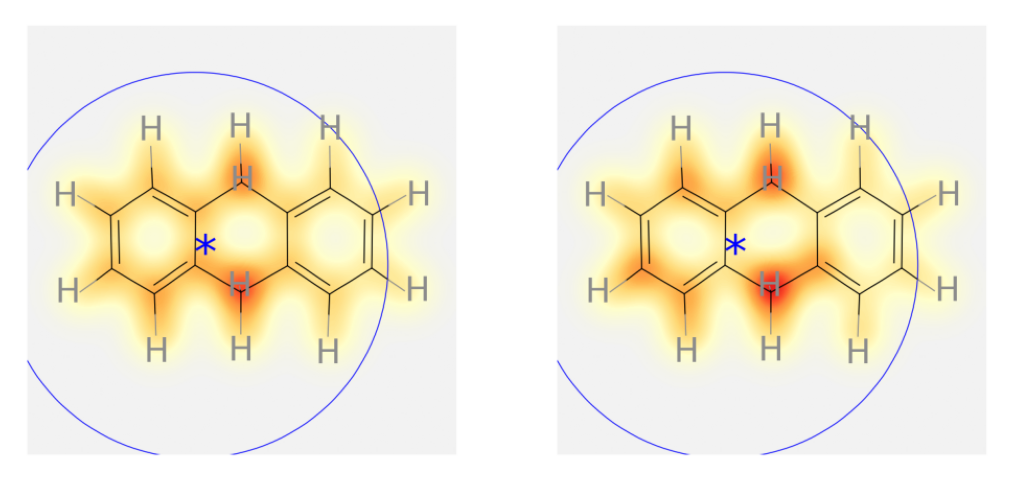}
        \label{fig:hlumo.dihydro.h}
    \end{subfigure}

    \caption{More examples of the comparison of GeoAttention between from $\text{GeoT}_{\textit{LUMO}}$ (left) and $\text{GeoT}_{\textit{H}}$ (right), respectively.}
    \label{fig:app.hlumo}
\end{figure}

\begin{figure}
\centering
    \begin{subfigure}{.48\textwidth}
        \centering
        \includegraphics[width=.95\linewidth]{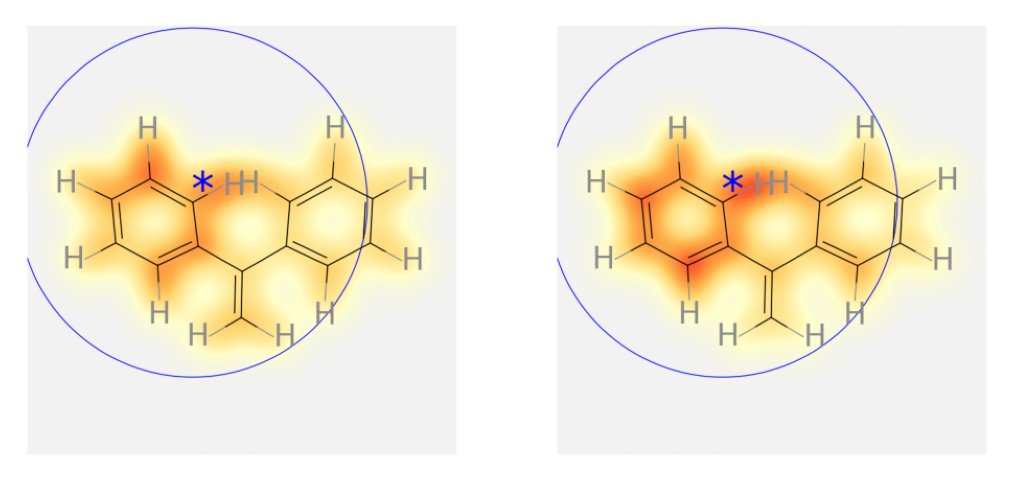}
        \label{fig:app.fig1.2.0}
    \end{subfigure}
    \hfill
    \begin{subfigure}{.48\textwidth}
        \centering
        \includegraphics[width=.95\linewidth]{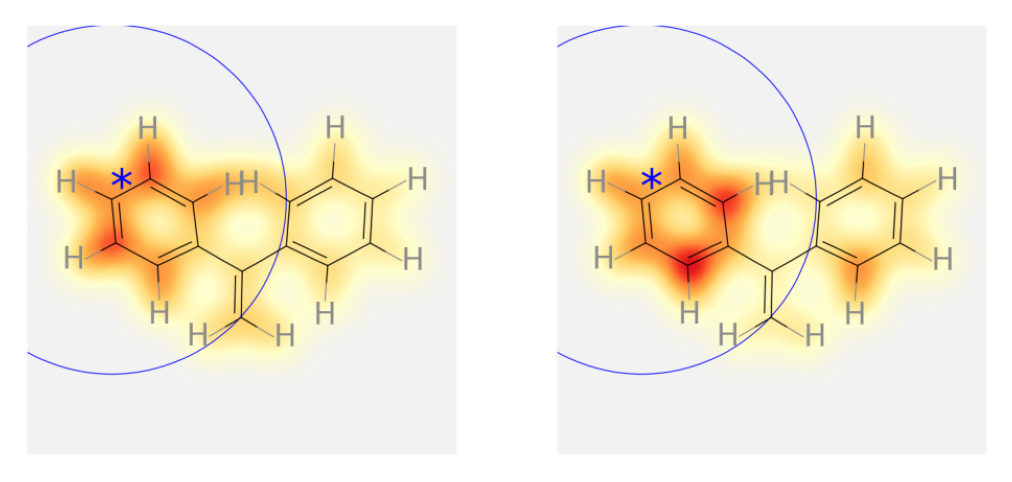}
        \label{fig:app.fig1.2.2}
    \end{subfigure}

    \medskip
    
    \begin{subfigure}{.48\textwidth}
        \centering
        \includegraphics[width=.95\linewidth]{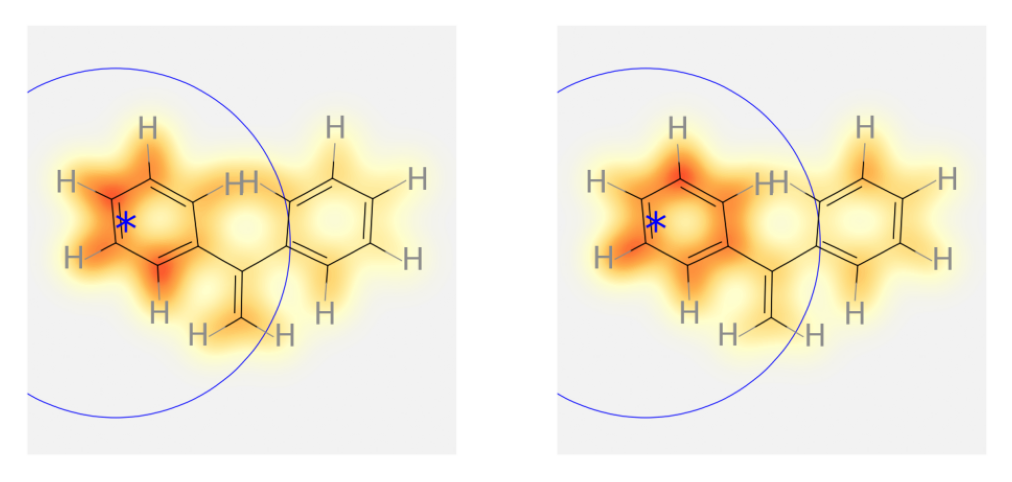}
        \label{fig:app.fig1.2.3}
    \end{subfigure}
    \hfill
    \begin{subfigure}{.48\textwidth}
        \centering
        \includegraphics[width=.95\linewidth]{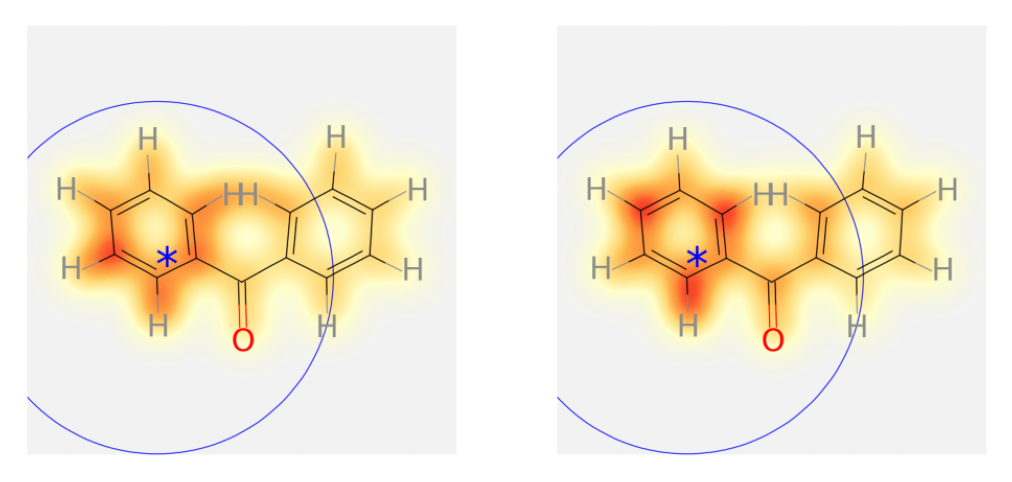}
        \label{ffig:app.fig1.3.0}
    \end{subfigure}
    
    \medskip

    \begin{subfigure}{.48\textwidth}
        \centering
        \includegraphics[width=.95\linewidth]{geometric_003_custom_0.png}
        \label{fig:app.fig1.3.2}
    \end{subfigure}
    \hfill
    \begin{subfigure}{.48\textwidth}
        \centering
        \includegraphics[width=.95\linewidth]{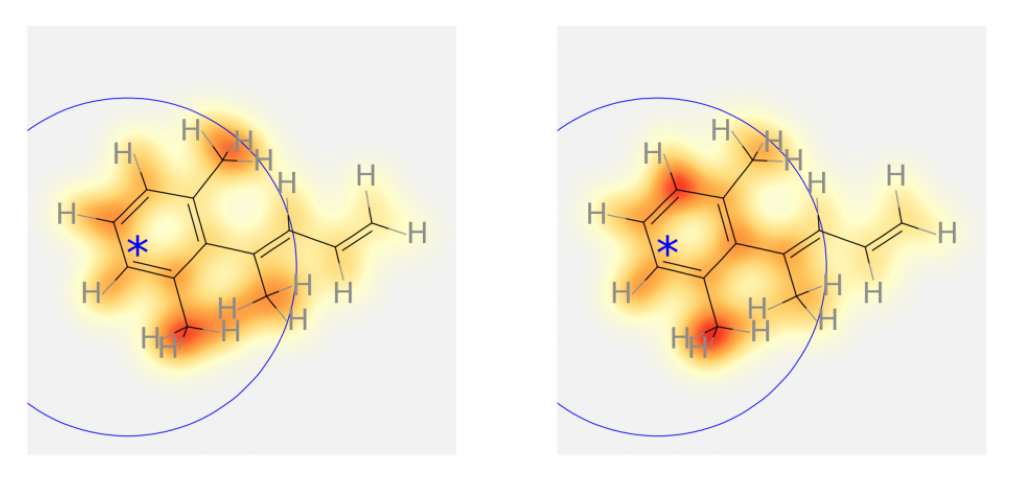}
        \label{fig:app.fig1.4.2}
    \end{subfigure}

    \medskip
    
    \begin{subfigure}{.48\textwidth}
        \centering
        \includegraphics[width=.95\linewidth]{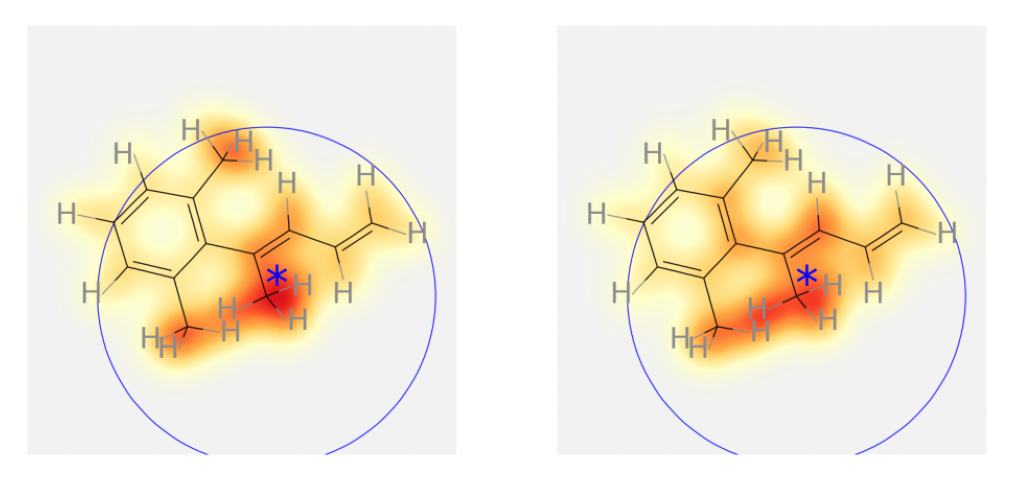}
        \label{fig:app.fig1.4.9}
    \end{subfigure}
    \hfill
    \begin{subfigure}{.48\textwidth}
        \centering
        \includegraphics[width=.95\linewidth]{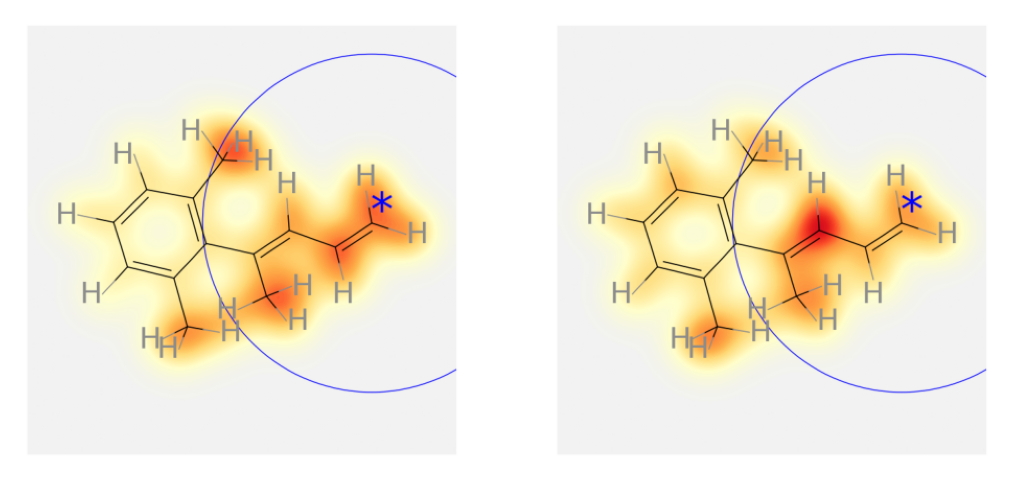}
        \label{fig:app.fig1.4.12}
    \end{subfigure}
    
    \caption{More examples of attention visualizations with aromatic rings. GeoAttention considers distance between atoms as well as conjugated regions. It shows that $\text{GeoT}_{\textit{LUMO}}$ can capture $\pi$-character of bonding mechanism.}
    \label{fig:app.fig1}
\end{figure}

\begin{figure}
\centering
    \begin{subfigure}{.21\textwidth}
        \centering
        \includegraphics[width=.95\linewidth]{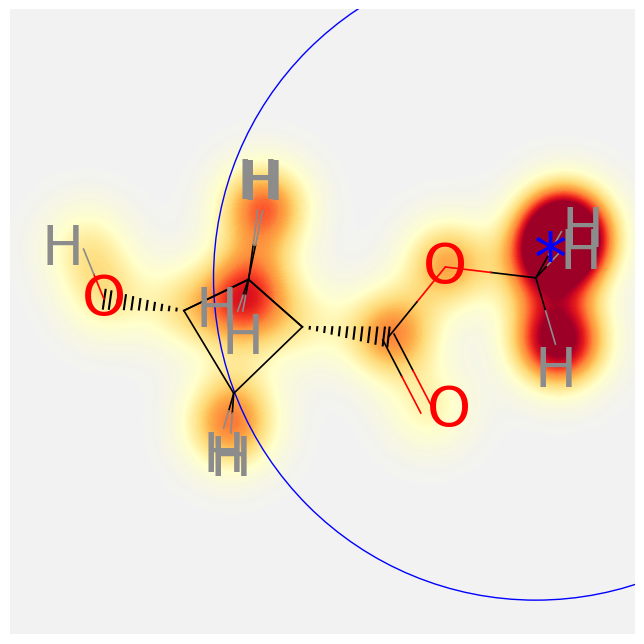}
    \end{subfigure}
    \hfill
    \begin{subfigure}{.21\textwidth}
        \centering
        \includegraphics[width=.95\linewidth]{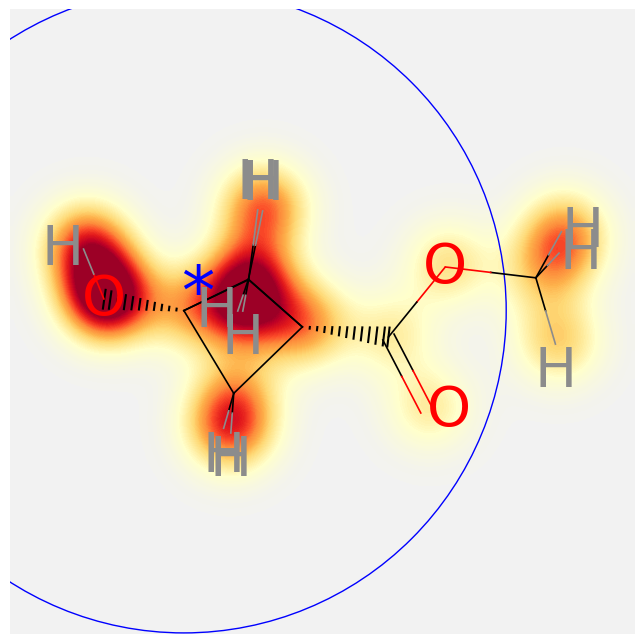}
    \end{subfigure}
    \hfill
    \begin{subfigure}{.21\textwidth}
        \centering
        \includegraphics[width=.95\linewidth]{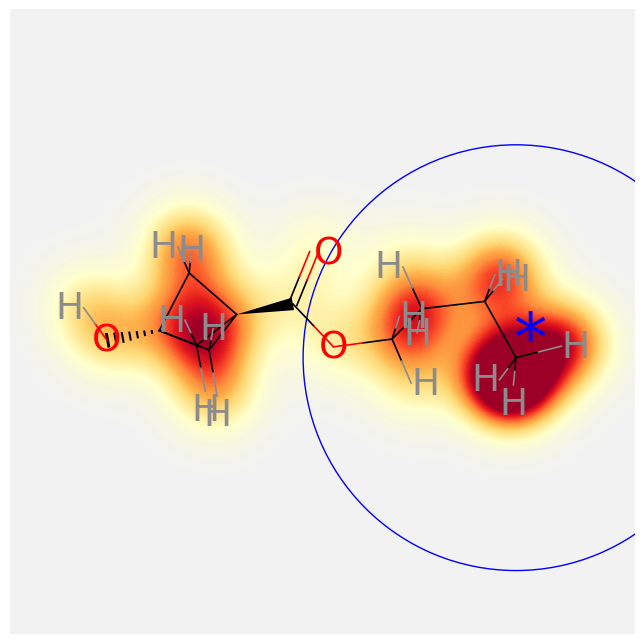}
    \end{subfigure}
    \hfill
    \begin{subfigure}{.21\textwidth}
        \centering
        \includegraphics[width=.95\linewidth]{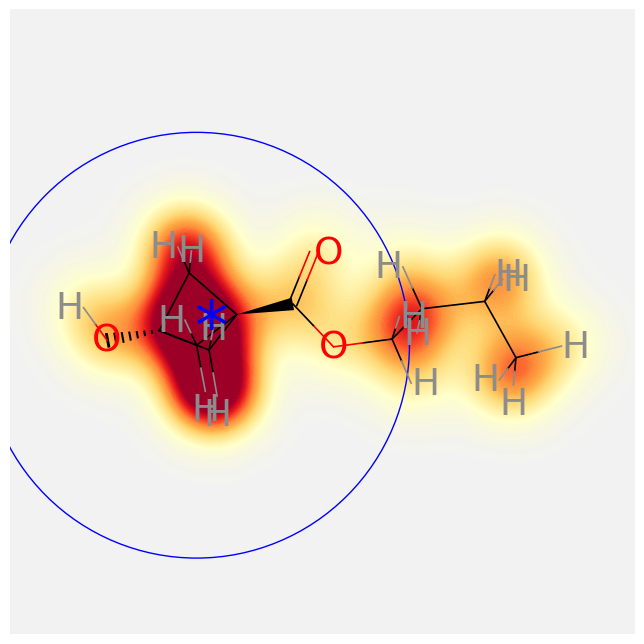}
    \end{subfigure}
    
    \medskip

    \begin{subfigure}{.21\textwidth}
        \centering
        \includegraphics[width=.95\linewidth]{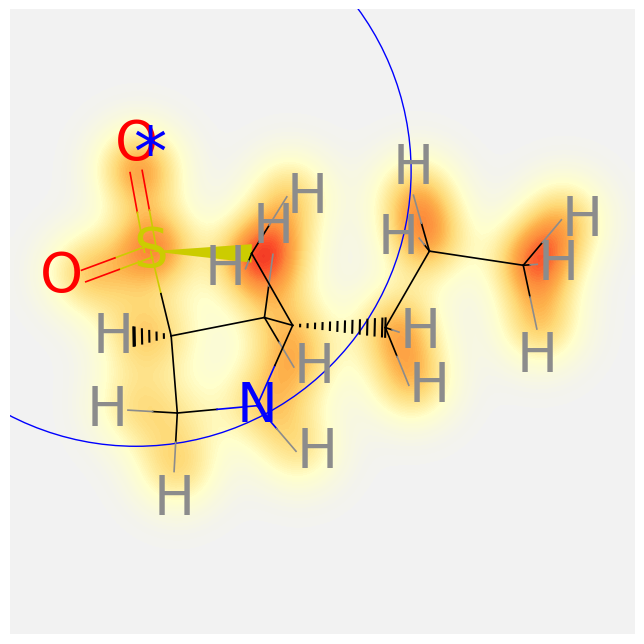}
    \end{subfigure}
    \hfill
    \begin{subfigure}{.21\textwidth}
        \centering
        \includegraphics[width=.95\linewidth]{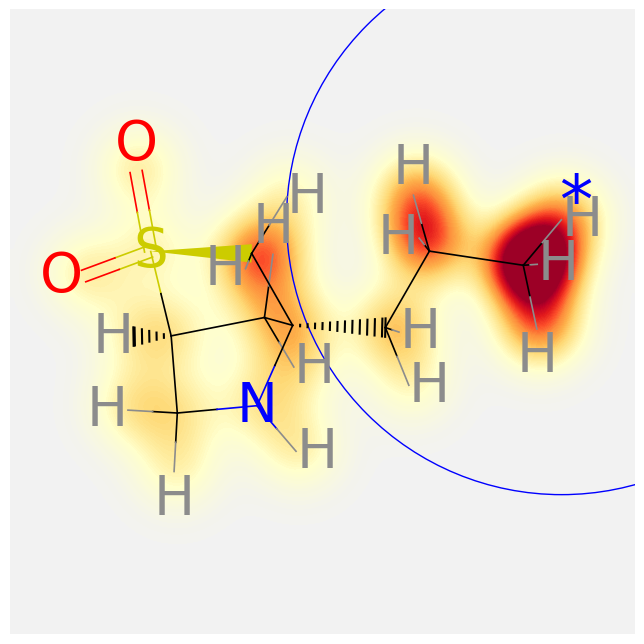}
    \end{subfigure}
    \hfill
    \begin{subfigure}{.21\textwidth}
        \centering
        \includegraphics[width=.95\linewidth]{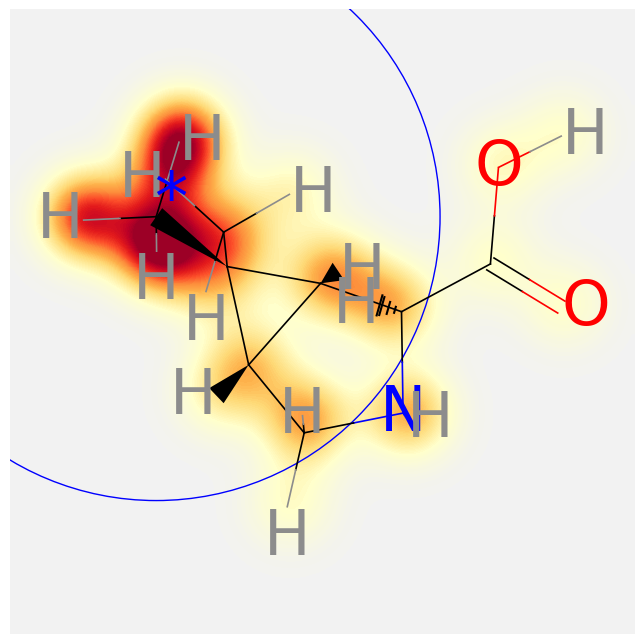}
    \end{subfigure}
    \hfill
    \begin{subfigure}{.21\textwidth}
        \centering
        \includegraphics[width=.95\linewidth]{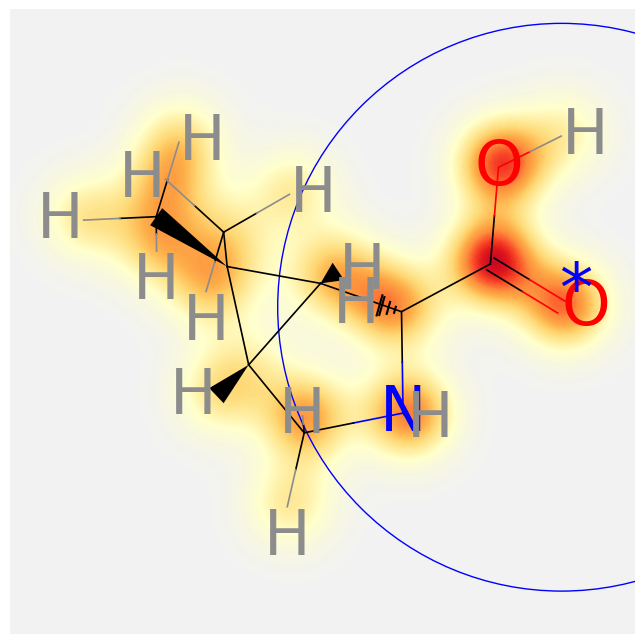}
    \end{subfigure}
    
    \medskip

    \begin{subfigure}{.21\textwidth}
        \centering
        \includegraphics[width=.95\linewidth]{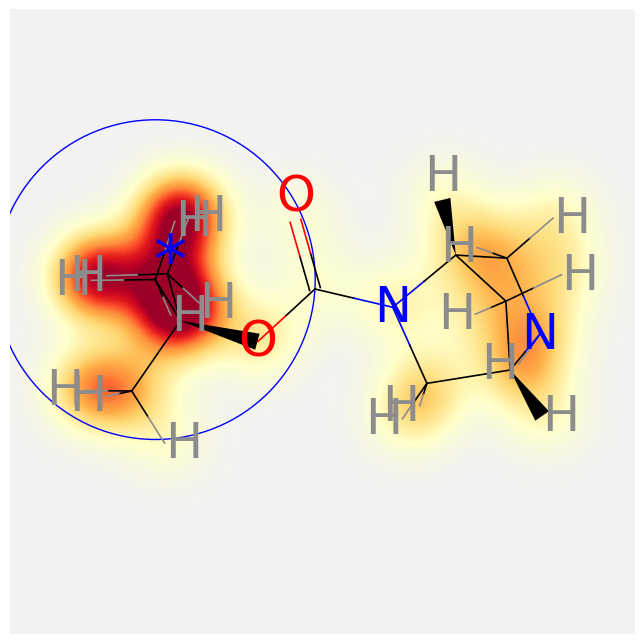}
    \end{subfigure}
    \hfill
    \begin{subfigure}{.21\textwidth}
        \centering
        \includegraphics[width=.95\linewidth]{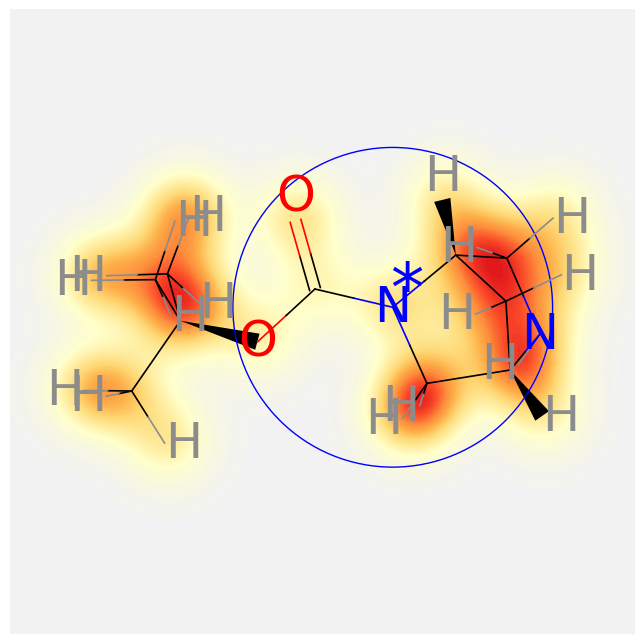}
    \end{subfigure}
    \hfill
    \begin{subfigure}{.21\textwidth}
        \centering
        \includegraphics[width=.95\linewidth]{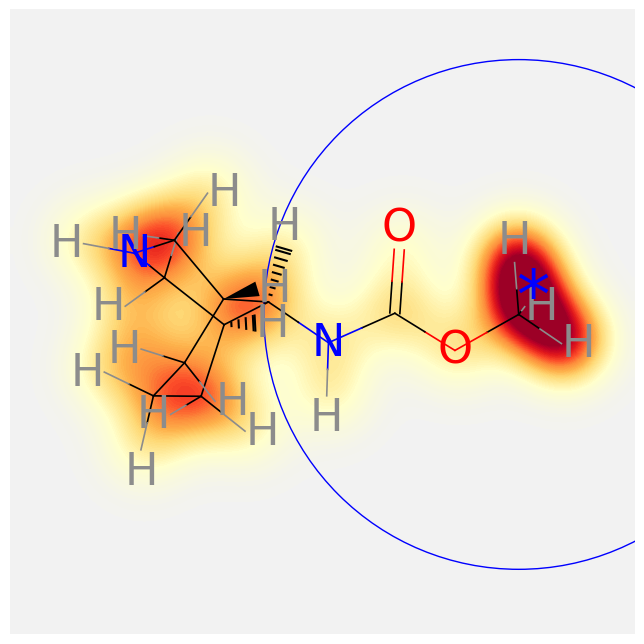}
    \end{subfigure}
    \hfill
    \begin{subfigure}{.21\textwidth}
        \centering
        \includegraphics[width=.95\linewidth]{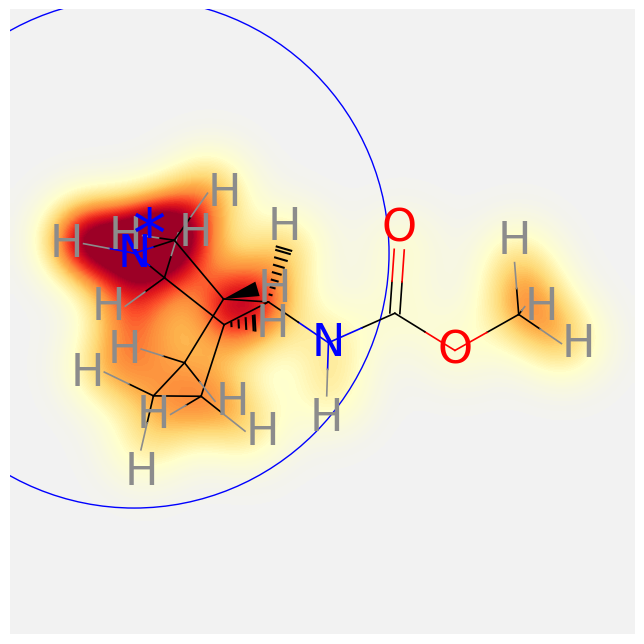}
    \end{subfigure}
    
    \medskip

    \begin{subfigure}{.21\textwidth}
        \centering
        \includegraphics[width=.95\linewidth]{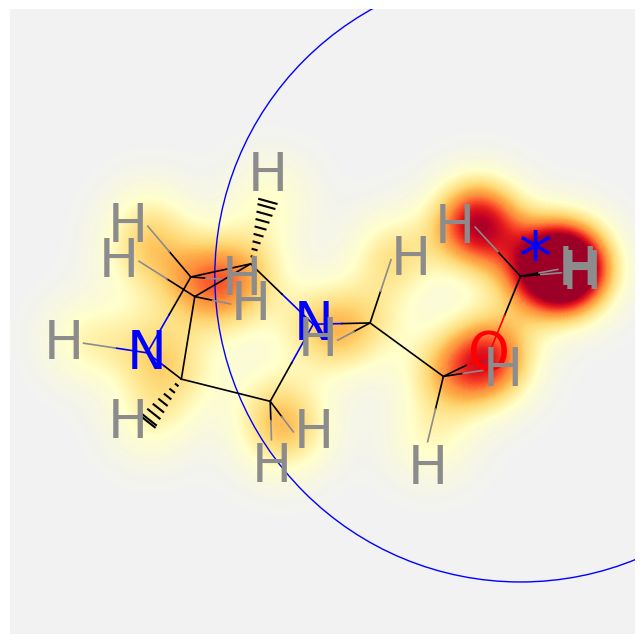}
    \end{subfigure}
    \hfill
    \begin{subfigure}{.21\textwidth}
        \centering
        \includegraphics[width=.95\linewidth]{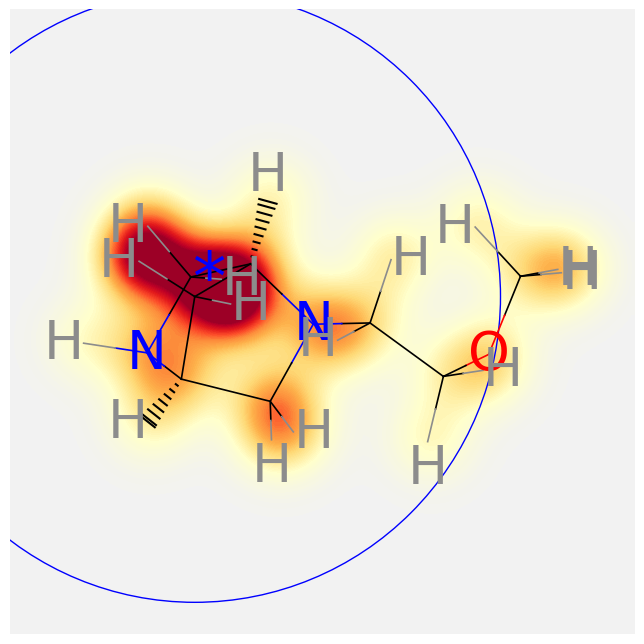}
    \end{subfigure}
    \hfill
    \begin{subfigure}{.21\textwidth}
        \centering
        \includegraphics[width=.95\linewidth]{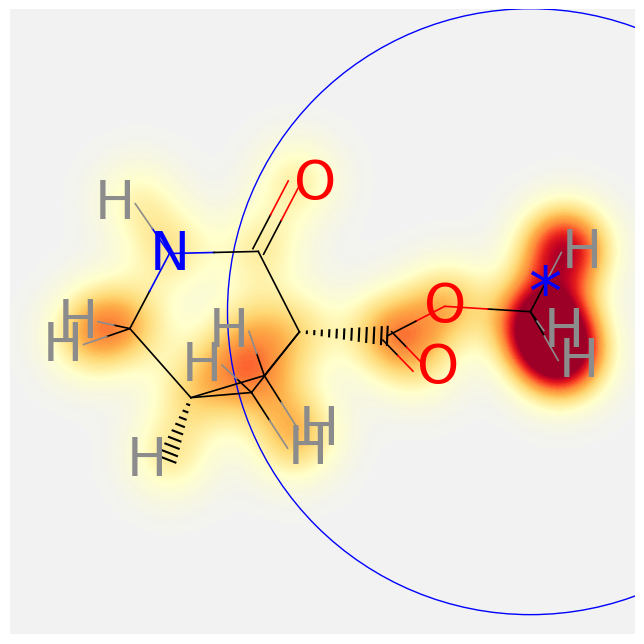}
    \end{subfigure}
    \hfill
    \begin{subfigure}{.21\textwidth}
        \centering
        \includegraphics[width=.95\linewidth]{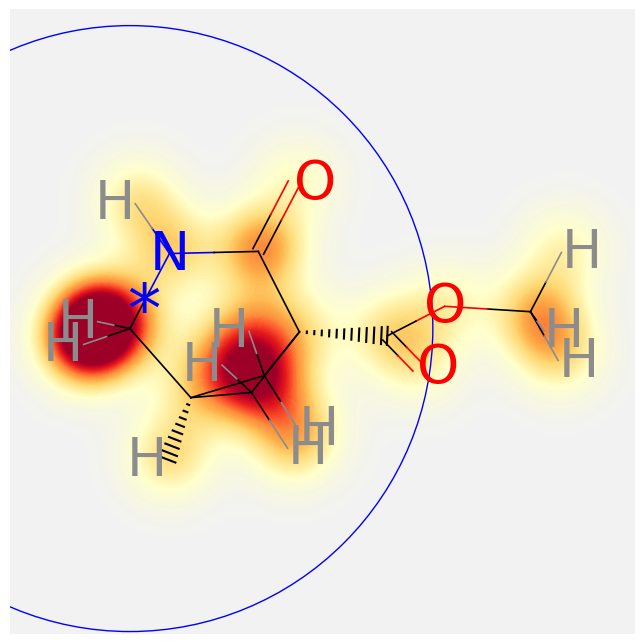}
    \end{subfigure}

    \caption{More examples of attention visualizations with high-strain rings. GeoAttention obtained from $\text{GeoT}_{\textit{H}}$ considers distance between atoms as well as steric effect induced by high-strain.}
    \label{fig:app.fig2}
\end{figure}

\section{Additional results of the model performances}
\label{Appendix.tables}

\begin{table}[ht]
  \caption{Mean absolute error on QM9.}
  \centering
  \label{tbl:mae_qm9}
  \begin{tabular}{llccc|cc|c}
    \toprule
    Target & Unit & SchNet & PhysNet & GRAT & Cormorant & DimeNet++ & GeoT \\
    \midrule
    $\mu$ & D & 0.033 & 0.0529 & 0.03898 & 0.038 & 0.0297 & 0.0297$^{*}$ \\
    $\alpha$ & ${a_0}^3$ & 0.235 & 0.0615 &0.07219 & 0.085 & 0.0435 & 0.05269 \\
    $\epsilon_{\mathrm{HOMO}}$ & eV & 0.041 & 0.0329 & 0.02228 & 0.034 & 0.0246 & 0.0250$^{*}$ \\
    $\epsilon_{\mathrm{LUMO}}$ & eV & 0.034 & 0.0247 & 0.02053 & 0.038 & 0.0195 & 0.0202$^{*}$ \\
    $\Delta_\epsilon$ & eV & 0.063 & 0.0425 & 0.035 & 0.061 & 0.0326 &  0.04392 \\
    <$R^2$> & ${a_0}^2$ & 0.073 & 0.765 & 0.76681 & 0.961 & 0.331 &  0.3008 \\
    $zpve$ & eV & 0.0017 & 0.0014 & 0.00208 & 0.0020 & 0.00121 & 0.00173$^{*}$ \\
    $U_0$ & eV & 0.014 & 0.0082 & 0.0705 & 0.022 & 0.00632 & 0.0111 \\
    $U$ & eV & 0.019 & 0.0083 & 0.02825 & 0.021 & 0.0062 & 0.0117 \\
    $H$ & eV & 0.014 & 0.0084 & 0.02549 & 0.023 & 0.0065 & 0.0113 \\
    $G$ & eV & 0.014 & 0.0094 &0.02505 & 0.020 & 0.0075 &  0.0117 \\
    $C_v$ & cal/mol K & 0.033 & 0.028 & 0.0326 & 0.026 & 0.023 & 0.0276 \\
    \bottomrule
  \end{tabular}
\end{table}

\begin{table}[ht]
  \caption{In-distribution validation energy MAE for OC20 IS2RE 10k task}
  \centering
  \begin{tabular}{lcl}
    \toprule
    model	& MAE \\
    \midrule
    
    SchNet*	& 1.059 \\
    SchNet (without PBC) & 1.077 \\
    DimeNet* &	1.0117 \\
    CGCNN*	& 0.9881 \\
    DimeNet++*	& 0.8837 \\
    \midrule
    GeoT	& 0.9631 \\
    GeoT + RBF	& 0.996 \\
    GeoT + ParallelMLP	& 	0.966 \\
    GeoT + AttnScale	& 1.018 \\
    GeoT + ParallelMLP	+ Attnscale & 	1.0516 \\
    GeoT + RBF + ParallelMLP + Attnscale  & 	0.9786 \\
    \bottomrule
  \end{tabular}
  \begin{tablenotes}
    \small
    \centering
    \item The results with * were retrieved from \cite{ocp_dataset}. Note that we did not implement PBC into GeoT, while other methods adopted it. PBC is a kind of a trick wherein the representation of input chemical is considered as the smallest unit of repeated structures. 
  \end{tablenotes}
    \label{tab:oc2010k}
\end{table}

\begin{table}[ht]
    \caption{In-distribution validation energy MAE for OC20 IS2RE full task.}
    \centering
    \begin{tabular}{lcl}
        \toprule
         model & MAE \\
         \midrule
         SchNet & 0.6458 \\
         SchNet (without PBC) & 0.729 \\
         \midrule
         GeoT (without PBC) & 0.7002 \\
         \bottomrule
    \end{tabular}
    
    \label{tab:oc20full}
\end{table}

\begin{table}[ht]
  \caption{Radial basis functions for ablation study.}
  \label{basis functions}
  \centering
  \begin{tabular}{lll}
    \toprule
    Basis         & Expression\\
    \midrule
    Linear     & $a + b \cdot d_{ij}$ &  \\
    Gaussian & exp($-\gamma||d_{ij} - d_{k}||^{2}$) \\
    Bessel   &  $\sqrt{\frac{2}{c}}\frac{\sin(\frac{n\pi}{c}d)}{d}$ \\
    \bottomrule
  \end{tabular}
\end{table}

\begin{table}[ht]
  \caption{Ablation study for linear, Gaussian, and Bessel basis functions in Table \ref{basis functions} in QM9 dataset.}
  \label{tbl:mae_ablation}
  \centering
  \begin{tabular}{llcccccccc}
    \hline
    Target &	Unit &	Gaussian basis &	Linear basis &	Bessel basis \\
    \hline
    $\mu$ &	D &	0.0299 &	0.664 &	0.0876 \\
    $\alpha$ &	${a_0}^3$ &	0.05269 &	0.752 &	0.15 \\
    $\epsilon_{\mathrm{HOMO}}$ &	eV &	0.02675 &	0.0659 &	0.056 \\
    $\epsilon_{\mathrm{LUMO}}$ &	eV &	0.02145 &	0.122 &	0.0326 \\
    $\Delta_\epsilon$ &	eV &	0.04392	 & 0.158 &	0.0575 \\
    <$R^2$> &	${a_0}^2$ &	0.3008 &	55.4 &	0.3262 \\
    $zpve$ &	eV	 & 0.00174 &	0.00845 &	0.00255 \\
    $U_0$ &	eV	 & 0.0111 &	0.2026 &	0.0147 \\
    $U$ &	eV	 & 0.0117 &	51.99 &	0.0275 \\
    $H$ &	eV &	0.0113 &	0.952 &	0.06 \\
    $G$ &	eV &	0.0117 &	0.574 &	0.013 \\
    $C_v$ &	cal/mol K &	0.0276 &	0.195 &	0.06 \\
    \hline 
  \end{tabular}
\end{table}

\section{The brief introduction of localized and delocalized interactions between atoms in molecules}
\label{Appendix.interactions}

Atoms are composed of a positively charged nucleus and negatively charged surrounding electrons. According to quantum mechanics, tiny particles such as electrons are inherently delocalized in space, and they form electron clouds. This diffusivity allows atoms to share electrons to make a chemical bond, and multiple bond-formation process eventually builds a molecule. Conversely, atoms in molecules can interact to each other by sharing electrons, and the type of atom-atom interactions can be classified by how the shared electrons are distributed in space. If shared electrons are confined between two atoms, the interaction is localized. If electrons are shared by more than two atoms, the interaction is considered to be delocalized.
When predict global molecular properties, the importance of delocalized interactions is obvious because the corresponding electron clouds could be spread over the entire molecular scaffold. However, successive localized interactions between atoms impose a certain geometry to a molecule and define specific bond angles and dihedral angles. This geometric constraint sometimes has great impact on the prediction of global molecular properties because it defines the spatial relationship between atoms that are not closely located to each other. In this section, we will describe how the localized and delocalized interactions are formed, and how they impact the global molecular properties that are the target of the GeoT prediction tasks.

\subsection{Localized interactions}
\label{appendix:local}

When two atoms are placed in close proximity, the electrons from each atom start to feel the attractive force exerted by the nucleus from the other atom. As a result, electrons are more densely confined between the inter-atomic space, and this localized strong interaction is called a $\sigma$-bond. If only two electrons are shared, then there can be only one $\sigma$-bond between two atoms. However, for some cases, more than two electrons can be shared between two atoms by making $\pi$-bonds.

In general, atoms in molecule are connected to multiple other atoms by chemical bonds. The number and the type of these localized bonds around a certain atom define the inter-atomic distances and bond angles around the atom. These constraints are imposed to each atom in molecule depending on the chemical environment surrounding them, and eventually restrict the possible conformation of a molecule.

The case could be analyzed in more detail by using graph theoretical expression. A graph representation of a molecule can be constructed by considering the atoms as nodes and connecting them if and only if the corresponding atoms are connected by chemical bonds. This graphical representation is useful because it captures the topology of molecules, but they are not capable of dealing with actual geometric parameters such as bond distances or angles. However, these parameters are indirectly accessible because 1-hop, 2-hop, and 3-hop interactions on this graph define bond distance, angle, and dihedral angle in 3-D space, respectively. Bond distance is a prime parameter for determining bond strength. Bond angle and dihedral angle also have a great impact of global molecular energy, as emphasized by angle strain of cyclic molecules or torsional strain of butanes. In fact, 4-hop or even the interactions at longer distances become important for conformational analysis of complex organic molecules, as exemplified by 1,3-diaxial interaction or syn-pentane interactions. It highlights that the global analysis for the successive localized interactions is crucial for the global molecular property prediction tasks.

\subsection{Delocalized interactions}
\label{appendix:delocal}

When multiple bond is formed between two atoms, the space around the inter-atomic axis is already occupied by electrons of $\sigma$-bond. Therefore, the electron clouds of $\pi$-bond are blown up to upper and lower space of the inter-atomic plane. If multiple $\pi$-bonds are connected by $\sigma$-bonds to form an alternating array of $\pi$- and $\sigma$-bonds, the diffuse lobes of $\pi$-bonds can be overlapped to each other. In this way, the corresponding electron clouds can be delocalized over multiple atoms. This special situation of chemical bonding is called $\pi$\textit{-conjugation}.

$\pi$-conjugation relies on long-range overlap between $\pi$-orbitals, and the degree of overlap is very sensitive to dihedral angle between them. Since the interaction is maximized at planar geometry, $\pi$-conjugation provides a molecule to have strong tendency to become planarized. In reality, this trend competes with other steric factors, as shown for the dihedral angle analysis of biphenyl molecule. When the $\pi$-conjugation makes a closed loop, the planar geometry is already enforced by the ring structure, and the stability of $\pi$-conjugation is maximized at this case. When the cycle is hexagon (or, in general, (4n+2)-gons), the system is called aromatic and shows exceptional stability because of other quantum mechanical reasons. It has been suggested that this aromaticity can facilitate the detection of delocalized interaction by a deep model \cite{vollhardt2010organic}.

\subsection{How localized and delocalized interactions affect molecular properties?}
\label{appendix:relationship}
In this paper, we deal with several molecular properties such as internal energy (U), enthalpy (H), free energy (G), and HOMO/LUMO energies. These properties can be classified into two categories. Molecular enthalpy (H) or free energy (G) reflect the overall energetics of a molecule, and mainly affected by nuclear-nuclear repulsion, electron-electron repulsion, and electron-nuclear attraction. These factors are mainly electrostatic, and thus strongly depend on bond distances between neighboring atoms. In other words, these properties are mainly determined by localized interactions.

On the other hand, HOMO or LUMO energies are dictated by delocalized interactions if such things exist in molecules. According to the quantum mechanical theory, the energy level of orbitals split when they are overlapped, and the degree of energy splitting is proportional to the degree of orbital overlap. Since orbitals involved in $\pi$-conjugations are overlapped by side-by-side manner, there overlap should be weaker than localized $\sigma$-bonds. Therefore, the molecular orbitals generated by $\pi$-conjugation are not too stabilized nor too destabilized. HOMO, the most unstable state among the bonding orbitals, and LUMO, the most stable state among the anti-bonding orbitals, are thus mainly generated by $\pi$-conjugation. Therefore, HOMO and LUMO energies tend to more rely on delocalized interactions rather than localized ones.

\end{appendices}


\end{document}